\documentclass{article}

\usepackage[preprint]{neurips_2023}

\usepackage[utf8]{inputenc} 
\usepackage[T1]{fontenc}    
\usepackage{hyperref}       
\usepackage{url}            
\usepackage{booktabs}       
\usepackage{amsfonts}       
\usepackage{nicefrac}       
\usepackage{microtype}      
\usepackage{amsmath}
\usepackage{multirow}

\usepackage{graphicx}
\usepackage{caption}
\usepackage{subcaption}
\usepackage{enumitem}
\usepackage{algorithm}
\usepackage{algpseudocode}
\usepackage[title]{appendix}
\usepackage{newtxmath}
\usepackage[table,xcdraw]{xcolor}
\usepackage{xspace}

\definecolor{red}{rgb}{1, 0, 0}

\newcommand{\ie}{\textit{i.e.}}
\newcommand{\eg}{\textit{e.g.}}

\makeatletter
\providecommand{\section}{}
\renewcommand{\section}{%
  \@startsection{section}{1}{\z@}%
                {-1.0ex \@plus -0.5ex \@minus -0.2ex}%
                { 1.0ex \@plus  0.3ex \@minus  0.2ex}%
                {\large\bf\raggedright}%
}
\providecommand{\subsection}{}
\renewcommand{\subsection}{%
  \@startsection{subsection}{2}{\z@}%
                {-0.75ex \@plus -0.5ex \@minus -0.2ex}%
                { 0.75ex \@plus  0.2ex}%
                {\normalsize\bf\raggedright}%
}
\providecommand{\subsubsection}{}
\renewcommand{\subsubsection}{%
  \@startsection{subsubsection}{3}{\z@}%
                {-0.5ex \@plus -0.5ex \@minus -0.2ex}%
                { 0.5ex \@plus  0.2ex}%
                {\normalsize\bf\raggedright}%
}
\providecommand{\paragraph}{}
\renewcommand{\paragraph}{%
  \@startsection{paragraph}{4}{\z@}%
                {0.3ex \@plus 0.2ex \@minus 0.2ex}%
                {-1em}%
                {\normalsize\bf}%
}
  
\makeatother

\title{Discrete, compositional, and symbolic representations through attractor dynamics}

\author{%
  Andrew J. Nam\thanks{Work done while at Mila and Stanford University}\\
  Princeton University \\
  Princeton, NJ \\
  \texttt{andrewnam@princeton.edu} \\
  \And
  Eric Elmoznino \\
  Mila, Université de Montréal \\
  Montréal, QC, Canada
   \\
  \texttt{eric.elmoznino@mila.quebec} \\
  \AND
  Nikolay Malkin\thanks{Work done while at Mila and Université de Montréal} \\
  University of Edinburgh \\
  Edinburgh, Scotland, UK \\
  \texttt{nmalkin@inf.ed.ac.uk} \\
  \And
  James L. McClelland \\
  Stanford University \\
  Stanford, CA \\
  \texttt{jlmcc@stanford.edu} \\
  \AND
  Yoshua Bengio\thanks{Senior author} \\
  Mila, Université de Montréal \\
  Montréal, QC, Canada \\
  \texttt{yoshua.bengio@mila.quebec} \\
  \And
  Guillaume Lajoie$^\dagger$ \\
  Mila, Université de Montréal \\
  Montréal, QC, Canada \\
  \texttt{guillaume.lajoie@mila.quebec}
}

\begin{document}

\maketitle

\begin{abstract}
Symbolic systems are powerful frameworks for modeling cognitive processes as they encapsulate the rules and relationships fundamental to many aspects of human reasoning and behavior. 
Central to these models are systematicity, compositionality, and productivity, making them invaluable in both cognitive science and artificial intelligence.
However, certain limitations remain. For instance, the integration of structured symbolic processes and latent sub-symbolic processes has been implemented at the computational level through fiat methods such as quantization or softmax sampling, which assume, rather than derive, the operations underpinning discretization and symbolicization.
In this work, we introduce a novel neural stochastic dynamical systems model that integrates attractor dynamics with symbolic representations to model cognitive processes akin to the probabilistic language of thought (PLoT).
Our model segments the continuous representational space into discrete basins, with attractor states corresponding to symbolic sequences, that reflect the semanticity and compositionality characteristic of symbolic systems through unsupervised learning, rather than relying on pre-defined primitives. 
Moreover, like PLoT, our model learns to sample a diverse distribution of attractor states that reflect the mutual information between the input data and the symbolic encodings. 
This approach establishes a unified framework that integrates both symbolic and sub-symbolic processing through neural dynamics, a neuro-plausible substrate with proven expressivity in AI, offering a more comprehensive model that mirrors the complex duality of cognitive operations.
\end{abstract}

\section{Introduction}

Symbols are ubiquitous in human activity and cognition, such as in language and mathematics.
However, unlike true symbolic systems such as computers, human cognition is supported entirely by neural connections.
While the emergence of symbolic thought from neural connections has been a major source of debate for several decades in cognitive science \citep{rumelhart1986parallel, fodor1988connectionism}, the recent development of large language models (LLMs) offers strong evidence that complex networks of neural connections can indeed perform symbol manipulation.
However, despite their apparent mastery of natural language, even the most sophisticated LLMs continue to struggle on symbolic problems, such as mathematics and reasoning \citep{achiam2023gpt, claude3model}.

Early critics of connectionist models \citep{fodor1988connectionism, marcus2003algebraic} have argued that this stems from a fundamental limitation of sub-symbolic models--those that utilize continuous, distributed forms of data representation, rather than discrete, clearly defined symbols--to study human cognition, instead opting for explicitly symbolic systems.
Newell, for instance, theorized that human thought could be simulated \citep{newell1980physical} using a computer program called the General Problem Solver (GPS) \citep{newell1961gps, newell1961computer} using a graph search algorithm.
Fodor formalized the philosophy into the language of thought (LoT) hypothesis \citep{fodor1988connectionism, fodor1975language, fodor2008lot, schneider2011language, pinker2003language} and proposed that at the core of human cognition is a latent set of rules and symbols with the systematic and compositional structure of language. 

In its more modern form, the probabilistic language of thought (PLoT) \citep{goodman2015concepts} extends the LoT framework through Bayesian statistics that allow for graded behavior consistent with human cognition that rigid rule-based systems struggle to capture.
Crucially, PLoT models capture \textit{distributions} over representations and hypotheses rather than making a single prediction, making them highly suitable for describing diverse, noisy, and multimodal data common in psychology.
Probabilistic models have been used to study a wide range of cognitive processes, including rule induction \citep{goodman2008rational}, cardinal principality acquisition \citep{piantadosi2012bootstrapping}, structural form discovery \citep{kemp2008discovery}, pragmatic language \citep{goodman2016pragmatic, degen2023rational}, and causal reasoning \citep{goodman2011learning, gopnik2004theory}.

LoT models focus on the behavior that a cognitive phenomenon produces, providing insights at Marr's computational level of analysis \citep{marr1982vision, griffiths2010probabilistic}, and are effective for understanding processes that can be described using symbols and programs.
Its key advantage is the compositional structure afforded by the combination of symbolic primitives, much like natural language, that facilitates the production of highly complex ideas using the rules that define the language and enables robust generalization to novel inputs beyond the training distribution.
However, specifying a LoT requires an explicit, predetermined set of symbols and processes that limits it to domains that can be symbolically articulated while abstracting away the sub-symbolic mechanisms underpinning these representations and computations \citep{mcclelland2010letting}--mechanisms that capture the many ways in which actual human reasoning and language processing deviate from purely symbolic computational systems.
These advantages and disadvantages have inspired the development of neurosymbolic hybrid models, combining both symbolic and sub-symbolic systems.



One such approach involves symbol production that is guided by a neural model, such as a language model that generates a program or intervening tokens, which serve as latent symbolic variables \citep{reed2015neural, ellis2018learning, ellis2023dreamcoder, chen2021evaluating, wei2022chain}.
Another approach uses neural networks as inference machines over symbolic variables, requiring approximations like slot attention mechanisms \citep{santoro2017simple, vaswani2017attention, locatello2020object}, quantization \citep{van2017neural, jang2016categorical, maddison2016concrete}, and other strategies to manage the continuous-discrete interface \citep{chaabouni2019anti, chaabouni2020compositionality}.
Among this latter approach is the GFN-EM algorithm \citep{hu2023gfnem},  which this paper builds on (see Section~\ref{sec:training}), that performs amortized inference over symbolic variables with neural networks.


Although these methods do allow symbolic representations in neural networks, optimization tricks to neurally approximate symbolic computation do not give a satisfying account of how symbols emerge from distributed neural activity.
First, vector quantization \citep{van2017neural} and the Gumbel-Softmax \citep{jang2016categorical, maddison2016concrete}, two popular mechanisms for imposing discrete bottlenecks in neural networks, assume a globally accessible array of state vectors in discrete slots, thereby building discreteness directly into the model architecture.
While this may be an effective module for artificial intelligence, it is unlikely that the human mind maintains explicit slots containing mental states, making this undesirable for an algorithmic account of human cognition.
Second, there is no account of how discretization occurs; it simply \textit{happens} by inheriting the values of the quantized state.
Consequently, this faces the same limitation as purely symbolic models: it defines the goal (discretization) at the computational level but fails to capture the underlying mechanism at the algorithmic level.
Third, people often exhibit traces of the original stimulus that decay over time, even when discrete-like behavior is observed \citep{mcmurray2002gradient}, which cannot be accounted for in models that simply map onto a discrete state.
Finally, these models provide no mechanism that would direct them to learn principled distributions of diverse discrete states that are characteristic of Bayesian statistics in PLoT models and human data, potentially resulting in more arbitrary solutions not capturing uncertainty well.

The present work seeks to address these limitations using a neural stochastic dynamical systems model that implements discretization by sampling attractor states that encode symbolic representations \citep{ji2024sources}.
Our approach draws inspiration from neurological studies that suggest that the brain may achieve and maintain discrete states through attractor dynamics \citep{khona2022attractor}. 
Integrator neurons in area MT of rhesus monkeys, for instance, have been found to converge on two distinct states \citep{shadlen2001neural} that correspond to each action in 2-action forced choice (2AFC) tasks resembling the dynamics of a drift diffusion model \citep{ratcliff1978theory}.
Moreover, optogenetic interventions in the anterior laterial motor cortex (ALM) of mice during a 2AFC task have provided causal evidence of attractor basins and decision boundaries \citep{inagaki2019discrete}.
While the neural trajectories and the subsequent action were robust to small perturbations, large perturbations would shift the trajectory towards the other attractor, causing the mouse to make the action that it otherwise would not have made given the observed stimulus.

Much like VQ-VAE and the Gumbel-Softmax, our model imposes a discrete bottleneck in its encoding process, but with two crucial differences.
First, rather than selecting discrete objects directly, our model learns an energy landscape and samples from it so that the underlying mechanisms are entirely sub-symbolic and formulated as continuous processes.
Second, the model learns a complex Bayesian posterior distribution over the token sequences conditioned on the input item.
While this property is common to variational latent samplers such as VAEs, normalizing flow models \citep{rezende2015variational}, and diffusion models \citep{sohl2015deep, ho2020ddpm}, our model is uniquely able to do so with both continuous states and discrete symbols.

To build a computational model with neural dynamics that are consistent with the attractor-based discretization mechanisms observed in biological systems \citep{inagaki2019discrete, khona2022attractor}, we borrow training methods from machine learning so that our model, \textit{once trained}, implements PLoT by sampling discrete symbolic sequences that represent the attractor states induced by inputs.
These attractors, which emerge during training from the learning objectives, are sampled according to a Bayesian posterior and the symbolic codes have compositional and hierarchical relations where appropriate.
However, at its core, the model is a dynamical system with graded activations with its knowledge embedded in the transition function, \ie, how each neuron influences the activity of other neurons, without any explicit architectural constraints that enforce discretization such as slotted vectors.
Crucially, none of the symbolic structure, the latent semantics, nor the attractor dynamics are specified a priori in the model as is done in probabilistic models, but are all entirely learned through unsupervised learning.
Instead, these properties emerge under the pressure to emit discrete tokens as outputs, potentially reflective of a reliance on highly distinctive units in the environment or experience, such as phonological elements in spoken languages \citep{jakobson1956phonology}.
Thus, our work provides a more complete account of both symbolic compositionality and its sub-symbolic implementation in a neural system, thereby bridging the gap between the explicit discrete representations of probabilistic models at the computational level and the latent distributed representations of connectionist models at the algorithmic level.


We begin by formalizing our approach and describing our model in the following section.
We then provide evidence that the model successfully captures the properties of PLoT through attractor dynamics.
First, we show that attractor states form around sentence embeddings, thereby mapping onto discrete symbolic representations.
Second, we demonstrate that these attractors reflect properties consistent with symbolic processes, such as grounded semantics and compositionality.
Third, we show that our model produces a diverse set of samples that approximate the Bayesian posterior that reflects the mutual information between an observed input and a discretized representation.
We also compare our model to VQ-VAE and an emergent communication model \citep{mu2021emergent}, and show that our model trained using attractor dynamics performs comparably to explicitly symbolic models that implement discretization by selecting vectors from a lookup table.
Finally, we conclude by discussing the implications of our model for cognitive and neurological modeling, key limitations, and possible future directions.

\section{Model}

\newcommand{\estep}[1]{\hbox{#1} \hbox{(E-Step)}}
\newcommand{\mstep}[1]{\hbox{#1} \hbox{(M-Step)}}

\renewcommand{\arraystretch}{1.2}
\renewcommand{\tabcolsep}{3pt}

\begin{table}[ht]
\centering
\caption{Modules in the model. Note that only the input encoder and forward dynamics modules (first two rows) are necessary for discretization. Remaining modules are used during training and possibly model analysis. Modules optimized during the E-step and M-step are parameterized by $\theta$ and $\phi$ respectively.}
\label{tab:attractors:modules}
\begin{tabular}{@{}p{0.22\linewidth}@{}p{0.55\linewidth}p{0.2\linewidth}@{}}
\toprule
\multicolumn{1}{c}{\textbf{Module}} & \multicolumn{1}{c}{\textbf{Description}} & \multicolumn{1}{c}{\textbf{Expression}} \\ \midrule
\mstep{Input  Encoder} & Encodes an object $x$ in the input space as an information-rich state $z_0$ in the latent representation space. & $P_\phi(z_0 \mid x)$ \\
\rowcolor[HTML]{EFEFEF} 
\mstep{Input  Decoder} & Decodes a state $z$ in the latent representation space to an object $x$ in the input space. & $P_\phi(x \mid z)$ \\
\estep{Forward  Dynamics} & Defines the forward stochastic transition dynamics from $z_t$ to $z_{t+1}$ in the latent representation space. & $P_\theta(z_{t+1} \mid z_t)$ \\
\rowcolor[HTML]{EFEFEF} 
\estep{Backward  Dynamics} & Defines the backward stochastic transition dynamics from $z_{t+1}$ to $z_t$ in the latent representation space. The backward dynamics are dependent on time $t$ and input $x$. & $P_\theta(z_t \mid z_{t+1}, t+1, x)$ \\
\mstep{Sentence  Encoder} & Encodes a token sequence $s$ as a state $\hat z_s$ in the latent representation space. & $ \hat z_s \leftarrow e_\phi(s)$ \\
\rowcolor[HTML]{EFEFEF} 
\estep{Sentence Decoder} & Decodes a state $z$ in the latent representation space to a token sequence $s$ autoregressively by sampling a token $s_i$ conditioned on $z$ and the previous tokens $s_{:i}$. We refer to the repeated application of the decoder to sample a full sequence $s \sim P_\theta(s \mid z)$ as the \textit{discretizer}. & $P_\theta(s_i \mid s_{:i}, z)$ \\
\estep{Flow  Correction} & Measures the flow for a state $z_t$ using a correction quantity $g$. & $g \leftarrow g_\theta(z_t, t, x)$ \\
\bottomrule
\end{tabular}
\end{table}

\renewcommand{\arraystretch}{1}
\renewcommand{\tabcolsep}{6pt}

\begin{figure}
\centering
  \includegraphics[width=\linewidth]{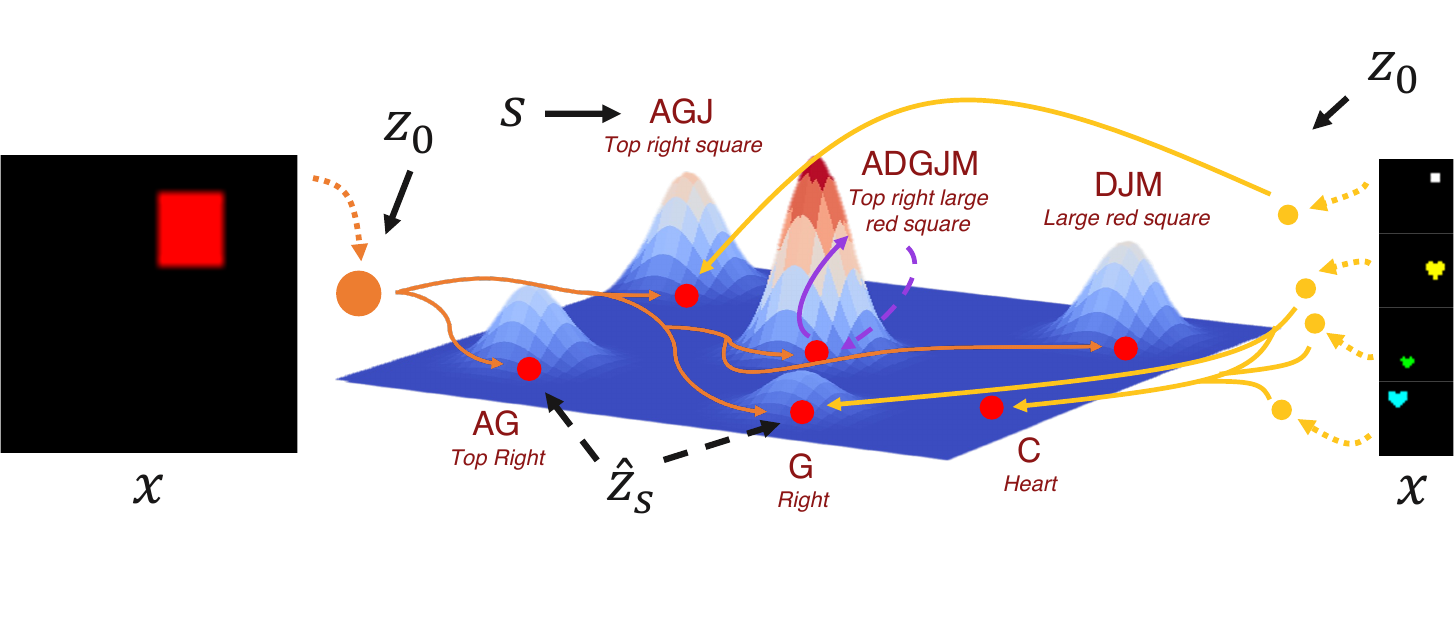}
\vspace{-1cm}
\caption{
    Model concept.
    Given an image of the red square as an input $x$, we first sample the maximally informative encoding $z_0 \sim P_\phi(z_0 \mid x)$.
    Starting at $z_0$ as the initial condition, we iteratively sample subsequent states according to the forward policy $z_{t+1} \sim P_\theta(z_{t+1} \mid z_t)$.
    The stochasticity allows the model to sample a diverse set of trajectories for a given input, while the attractor dynamics pull the trajectories towards sentence embeddings $\hat z_s$, so that the distribution of terminal states $P_\theta(z_T \mid x)$ (represented as the density in the figure) forms discrete attractor basins around the embeddings.
    Each sentence embedding $\hat z_s$ corresponds to a token sequence $s$, and the relative density around $\hat z_s$ corresponds to how well the sentence describes the red square image.
    In this example, the states around the embedding for the token sequence that translates to `top right large red square' is sampled most frequently, while states around the embedding representing `heart' are effectively never sampled.
    Other inputs (shown on the right) map to their own respective encodings $z_0$ and produce trajectories towards appropriate sentence embeddings, \eg, `heart' for the images containing hearts.
    Note that we show English words in this diagram for interpretability; the model uses arbitrary tokens with emergent semantics without exposure to any formal language.
}
\label{fig:attractors:concept}
\end{figure}

Our model consists of a collection of modules shown in Table~\ref{tab:attractors:modules}, including several auxiliary modules described in S\ref{sec:training}. We make a distinction between two categories of components based on their training step, explained below: the encoder/decoder modules whose parameters are denoted $\phi$, and the latent/dynamics modules governing sampling processes whose parameters are $\theta.$

\begin{figure}[t]
\centering
\begin{subfigure}{.49\linewidth}
  \centering
  \includegraphics[width=\linewidth]{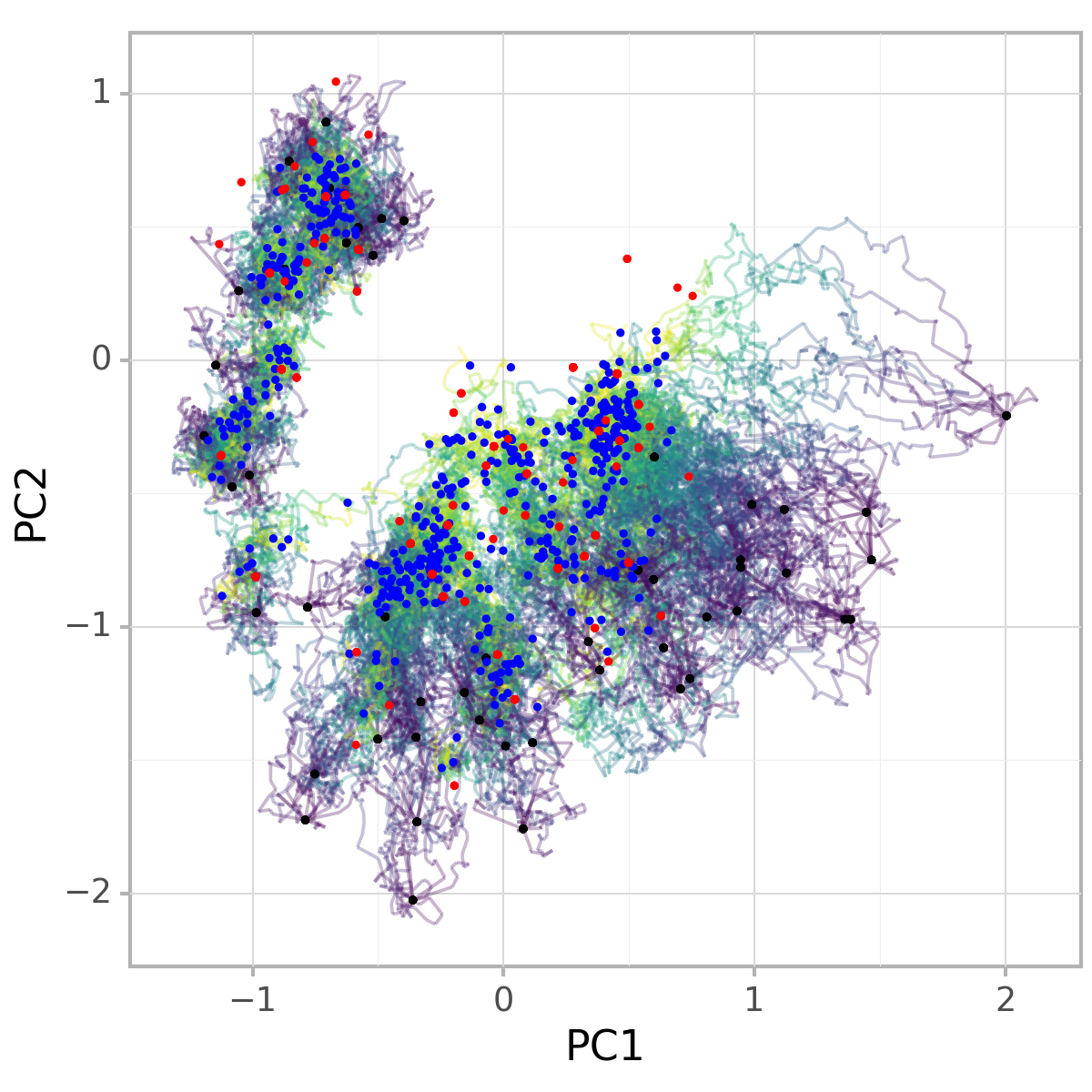}
  \caption{}
\end{subfigure}
\begin{subfigure}{.49\linewidth}
  \centering
  \includegraphics[width=\linewidth]{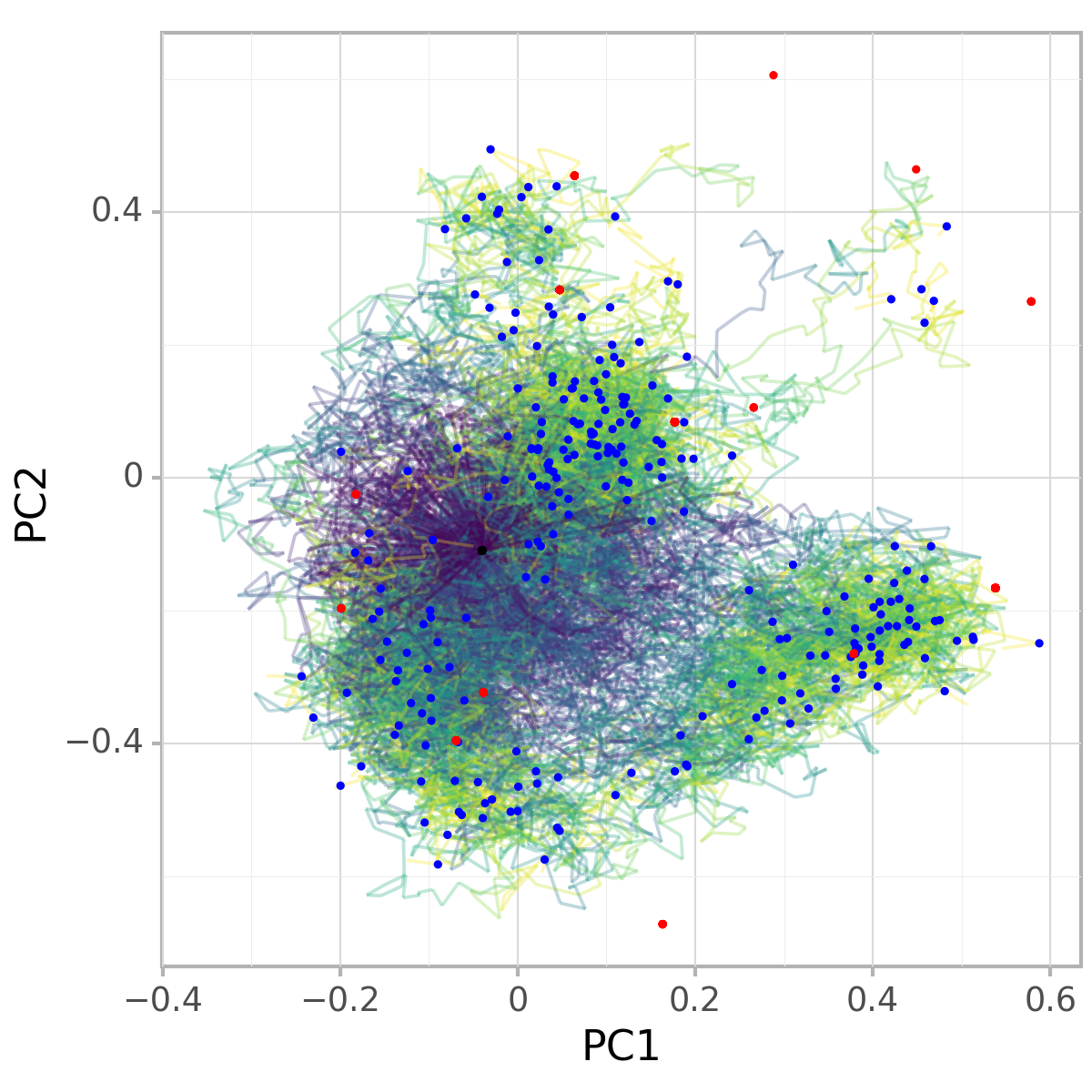}
  \caption{}
\end{subfigure}
\caption{
Projections of sample trajectories to two principal dimensions for the HBV dataset. Black dots indicate initial conditions $z_0$. Blue dots indicate terminal states $z_T$. Red dots indicate sentence embeddings $\hat z_w$.
(a) Trajectories sampled using 5 exemplars from 10 different classes.
(b) Trajectories sampled from a single exemplar.
Note that the two figures are scaled differently.
}
\label{fig:attractors:attractors}
\end{figure}

Our aim is to learn a marginal distribution $P_\theta(z_T \mid x)$ of distributed representations in latent space $\mathcal{Z}$ of an input $x \in \mathcal{X}$, \eg\ an image to be described, where the distribution is characterized by clusters around embedding vectors $\hat z_s$ of discrete token sequences $s \in \mathcal{S}$ proportional to how well $s$ describes $x$.
This encoding is performed gradually through a trajectory $z_0 \rightsquigarrow z_T = z_0, z_1, \ldots, z_T$ in $\mathcal{Z}$ over $T$ discrete time steps, starting at an input embedding $z_0$ that contains high mutual information with the input $x$, sampled from an input encoder module $P_\phi(z_0 \mid x)$.
From $z_0$, we iteratively sample the next point in the trajectory $z_{t+1}$ from a continuous-state stochastic forward dynamics module $P_\theta(z_{t+1} \mid z_t)$ until some convergence condition to get the latent trajectory $z_0 \rightsquigarrow z_T$.
$z_T$, which is expected to be near an attractor point that corresponds to the embedding vector $\hat z_s \in \mathcal{Z}$ of some $s$, may then be decoded into a discrete token sequence $s = s_1,\  \ldots, \ s_k,\ s_{eos}$ by starting with an empty sequence $s_{:1} = \varnothing$ and iteratively adding a token $s_i$ sampled from a sentence decoder module $P_\theta(s_i \mid s_{:i}, z_T)$ to the existing sequence $s_{:i}$ until an end-of-sequence token $s_{eos}$ is drawn.
Fig.~\ref{fig:attractors:concept} illustrates the model concept and the sampling process.
Fig.~\ref{fig:attractors:attractors} shows 2-D projections of sampled trajectories from a diverse set of inputs (a) and from a single input with multiple attractor modes (b).

The input encoder $P_\theta(z_T \mid x)$, the forward dynamics $P_\theta(z_{t+1} \mid z_t)$, and the backward dynamics $P_\theta(z_t \mid z_{t+1}, t+1, x)$ are all represented using conditional Gaussian distributions.
The means and variances of these distributions are determined by neural networks, effectively making the policy a discretized neural stochastic differential equation as described by \citep{tzen2019neural} and the sampling process equivalent to the Euler-Maruyama method \citep{KloedenPlaten1992}, a numerical technique used to simulate stochastic differential equations.
To model gradual changes in neural activity over time, we apply a prior over the magnitude of the state change using a bounded mean on the transition distribution.
The relationship between a sentence embedding $\hat z_s$ in continuous space and its tokenized form is represented using a stochastic discretizer function $s \sim P_\theta(s \mid z)$ and a deterministic embedding function $e_\phi: s \mapsto \hat z_s$
\footnote{We use $\theta$ and $\phi$ to denote the training step that they are optimized. See Section~\ref{sec:training}}.
The trajectory unfolding from $z_0$ to $z_T$, beginning with the initial encoding and culminating at the attractor point, exemplifies the continuous dynamics that facilitate the transformation of rich information into discrete, compositional, and stable thoughts.

The model is trained by considering that sampling from $P_\theta$ must be proportional to a learned unnormalized density which we call the reward function\footnote{This is related to the energy function $\mathcal{E}$ in energy-based models by $R = e^{-\mathcal{E}}$.}, $R_\phi(z_T, s; x)$ in Eq.~\ref{eq:reward}\footnote{$\phi$ is fixed when optimizing $\theta$.}:
\begin{equation}
\label{eq:reward}
    P_\theta(z_T, s \mid x) \propto 
    R_\phi(z_T, s; x) = \exp (-D_{\rm KL}( P_\phi(z_0 \mid x) \ \| \ P_\phi(z_0 \mid \hat z_s))) \cdot P_\phi(s \mid z_T) \cdot P(s).
\end{equation}
The above reward consists of several sub-objectives. First, the KL-divergence term $D_{\rm KL}( P_\phi(z_0 \mid x) \ \| \ P_\phi(z_0 \mid \hat z_s))$ measures the amount of information lost through discretization, so that its negation reflects the informational fidelity of the input in the tokenized form.
Next, $P_\phi(s \mid z_T)$ represents the likelihood of different sentences from the same embedding, independent of the object $x$ being described, which can offset synonymous sentences sharing similar embeddings. 
Intuitively, if one were to observe an apple and think `red', they could just as easily have thought `scarlet' without significantly altering the content of the thought (ignoring for the moment that `red' is more common than `scarlet').
Here, we make a simplifying assumption that they would have been no less likely to arrive at the latent representation for redness had they only known the word `red' and not `scarlet'.
Thus, we scale the reward function around the region in latent space such that a $z_T$ would be sampled no differently with just one nearby $\hat z_s$ of a symbolic sequence corresponding to it than with ten.
Lastly, $P(s)$ specifies a prior over $\mathcal{S}$, \eg, in the simplest case a Gricean preference over shorter sequences \citep{grice1975logic}.


\subsection{Training}
\label{sec:training}

We describe the essential components of the training algorithm here and direct the reader to Appendix~\ref{sec:training_details} for further details, including the training objective functions.
A summary of each individual module can be found in Table~\ref{tab:attractors:modules}.


We train our model as a continuous generative flow network (GFN) \citep{bengio2021foundations, lahlou2023theory}: a method for learning a generative model that samples a trajectory of states such that the distribution of terminal states is proportional to a given unnormalized target density. 
Our training procedure alternates steps in an expectation-maximization (EM) loop \citep{hu2023gfnem} in which the forward dynamics and discretizer modules serve as a posterior estimator.
The GFN-EM algorithm builds on the general EM algorithm \citep{dempster1977em}, a method to estimate the values of latent variables that are involved in generating observations.

Intuitively, the two steps of the GFN-EM algorithm work in the following way.
During the expectation step (E-step), the model assumes the semantics of each token sequence and aims to answer: 
\textit{Given the space of possible sentences and their meanings, which sentences should be used to describe each observation?}
Conversely, during the maximization step (M-step), the model assumes the sentences that have been used to describe each observation and aims to answer:
\textit{Given the observations and the sentences used to describe them, what should the sentences mean?}
Thus, the two steps work in conjunction, labeling observations with sentences during the E-step and determining the sentence semantics during the M-step.
We describe both steps in detail below.


\paragraph{E-step}
The E-step optimizes the parameters involved in the sampling policy, which we collectively denote using $\theta$, consisting of the forward dynamics $P_\theta(z_{t+1} \mid z_t)$ and the discretizer $P_\theta(s_i \mid s_{:i}, z)$.
The forward dynamics model is trained alongside an auxiliary backward dynamics model $P_\theta^{\leftarrow}(z_t \mid z_{t+1}, t+1, x)$ that learns the policy for sampling a trajectory in reverse.
The discretizer module $P_\theta(s \mid z)$ constructs the token sequence by adding a single token at a time, conditioned on the previous tokens and the latent state $z$, by sampling from $P_\theta (s_i \mid s_{:i}, z)$.
This module may be implemented without constructing the token sequences in a predefined order (\eg, by appending a token $s_i$ to the end of $s_{:i}$) by training an additional backward policy model $P^\leftarrow_\theta (s_i \mid s_{:i+1}, z, x)$ which predicts the sequence $s_{:i}$ before the $i$th token was added.
Since the reward function does not involve $z$, the discretizer also requires a model that predicts the state $z$ from which the sequence was sampled from: $P^\leftarrow_\theta(z \mid s, x)$.
However, because this is equivalent to the first backward step in the latent space $P(z \mid \hat z_s, x)$, it may computed using the backward dynamics model rather than as a separate module.

All parameters are jointly optimized using the detailed balance objective with forward-looking flow parametrization (FL-DB), which involves the flow correction module $g_\theta(z,x,t)$ \citep{bengio2021foundations,pan2023better} to estimate the `flow' of a state, a measure unique to generative flow network models that measures the expected future reward from a state $z_t$.
Intuitively, this training process aims to match the joint probability of a trajectory and token sequence $P(z_0, \ldots, z_T, s \mid x)$ between the forward sampling process (Eq.~\ref{eq:attractors:traj_prob_forward}) that begins with an input embedding $z_0$ 
\begin{equation}
\label{eq:attractors:traj_prob_forward}
P(z_0, \ldots, z_T, s \mid x) = P(z_0 \mid x) \cdot P(z_1 \mid z_0) \cdots P(z_T \mid z_{T-1}) \cdot P(s \mid z_T)
\end{equation}
and the backward sampling process (Eq.~\ref{eq:attractors:traj_prob_backward}) that begins with a token sequence $s$
\begin{equation}
\label{eq:attractors:traj_prob_backward}
P(z_0, \ldots, z_T, s \mid x) = P(s \mid x) \cdot P(z_T \mid \hat z_s, T+1, x) \cdot P(z_{T-1} \mid z_T, T, x) \cdots P(z_0 \mid z_1, 1, x).
\end{equation}


\paragraph{M-step} 
The M-step optimizes the parameters of modules involved in the reward function, which we collectively denote using $\phi$. 
The distribution over initial input embeddings $P_\phi(z_0 \mid x)$ and the distribution of objects described by $s$ in latent space $P_\phi(z_0 \mid \hat z_s)$ are both parameterized as Gaussian distributions where the means and variances are predicted by a neural network.
We also optimize an auxiliary reconstruction model $P_\phi(x|z_0)$ that promotes high mutual information between $x$ and the initial point of the dynamics $z_0$.

For each input $x$, we first sample an initial embedding $z_0 \sim P_\phi(z_0 \mid x)$, a trajectory $z_0 \rightsquigarrow z_T$ according to the dynamics model $P_\theta(z_{t+1} \mid z_t)$, and a symbol sequence from the terminal point $s \sim P_\theta(s \mid z_T, x)$.
Then, for a sampled tuple $(x, z_0, s)$, we optimize $P_\phi(z_0 \mid x)$ using a variational autoencoder \citep{Kingma2014AutoEncodingVB} with an additional KL-divergence term to increase the mutual information between $x$ and $s$.




\subsection{Role of explicit symbolic form}
We have noted the necessity of explicit symbolic forms $s$ during training.
Here, we expand on what roles they play in shaping the model.
First, the symbolic forms represent the set of stable concepts learned by the model with the constraint that each symbolic sequence maps onto a narrow region in the latent space, which we abstract as an embedding vector $\hat z_s$.
These embedding vectors implicitly define the set of initial conditions for the backward dynamics, which are subject to the same smoothness prior as the forward dynamics, and therefore form smooth trajectories that spread out with relatively small step sizes.
Thus, the backward policy forms repulsor states at the sentence embeddings that mirror attractor states for the forward policy.

Second, the symbolic forms provide the frame for supporting inductive biases towards systematicity and compositionality.
The shared model weights used in processing symbolic sequences naturally encourage semantic and syntactic consistency that can then propagate towards shaping the latent space with the same properties, resulting in an inductive bias that favors linguistic structure.
For instance, it is possible to explicitly define a prior $P(s)$ that places a cost for sequence complexity, such as the sequence length, which is easily defined using symbolic forms but is more  difficult using continuous vectors. In absence of these symbolic forms, the attractor dynamics on their own only place an inductive bias for discrete representations that need not have additional compositional structure.

Although these are important constraints for \textit{training} the model, we emphasize that the diffusion dynamics alone are sufficient to capture the mechanisms underlying discretization.
Thus, a \textit{trained} model may be interpreted fully sub-symbolically without the final tokenization step while maintaining the characteristics consistent with symbolic representations and processing.
We return to this distinction and limitation in the Discussion section.

\section{Experiments}

\begin{figure}[t]
\centering
\includegraphics[width=\linewidth]{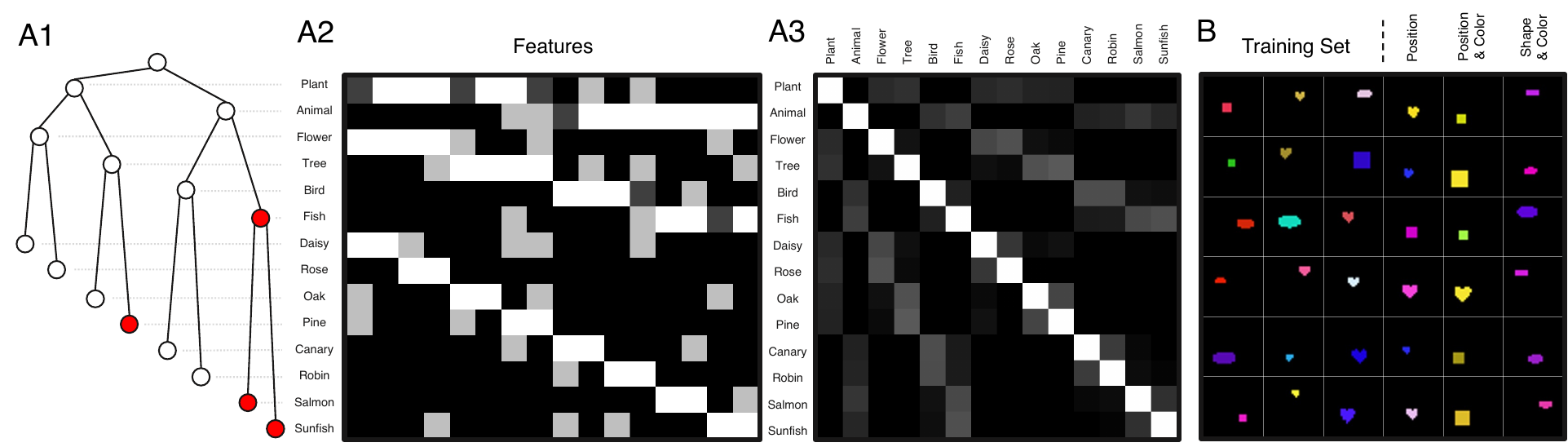}
\caption{
    (A) Simplified HBV dataset with 16-bit vectors and a depth-3 tree.
    (A1) The underlying hierarchical structure. Nodes in red indicate held-out classes.
    (A2) Prototype vectors and sample exemplars. White and black bits indicate where the prototype and sample are both 1 and 0 respectively. Light gray bits indicate where the sample is 1 but the prototype is 0. Dark gray bits indicate where the sample is 0 but the prototype is 1.
    (A3) Feature correlation matrix.
    (B) Samples from the dSprites dataset. Left three columns show samples from the training set and the right three columns from the held-out sets.
}
\label{fig:dataset}
\end{figure}


For our experiments, we used three-layer multilayer perceptrons (MLPs) to configure the dynamics models $P_\theta(z_{t+1} \mid z_t)$ and $P^\leftarrow_\theta(z_{t-1} \mid z_t, x, t)$, the discretizer models $P_\theta(s_{i+1} \mid s_{:i}, z, x)$ and $P^\leftarrow_\theta(s_i \mid s_{:i+1}, z, x)$, and flow models $g_\theta(z, x, t)$ and $Z_\theta(z, x)$.
The encoder $P_\phi(z_0 \mid x)$ and decoder $P_\phi(x \mid z_0)$ were first pre-trained as a variational autoencoder \citep{Kingma2014AutoEncodingVB}.
To focus our analyses, we introduced several simplifications to our model without compromising its efficacy: utilizing the mean of $P_\phi(z_0 \mid x)$ instead of drawing samples for $z_0$ at each iteration, substituting $z_0$ for $x$ in all other modules where $x$ is an input, employing a fixed backward policy for the first step from a sentence embedding $P^\leftarrow(z \mid \hat z_s, T+1, x) = \mathcal{N}(\hat z_s, \epsilon^2)$, simplifying the structure of our token sequences to a bag-of-words format, and using a uniform prior over $\mathcal{S}$.
We set the latent trajectory length to 20 steps, with extensions in certain analyses to ensure convergence.
Each analysis was performed using 5 instances of each model variant.

\subsection{Hierarchical binary vectors (HBV)}
To investigate the ability of our models to capture hierarchical structures effectively, we use a dataset inspired by real-world classifications, such as distinguishing initially between broad categories like plants and animals, and further subdividing into trees, flowers, fish, and birds (Fig.~\ref{fig:dataset}a).
We abstract this structure using $N$-bit binary vectors that mimic an idealized phylogenetic tree \citep{rogers2003semantic, saxe2019mathematical} with depth $(D)$, where each node represent a class prototype.
Starting with a root node of all \texttt{1}s, each node splits into two children that inherit non-overlapping halves (left and right) of their parent's \texttt{1}s (Fig.~\ref{fig:dataset}b).
This bifurcation continues recursively until the tree reaches its designated depth \(D\) with the deepest (leaf) prototypes containing $\frac{N}{2^D}$ \texttt{1}s.
To generate exemplars from a prototype, we independently sample the value of each bit (see Fig.~\ref{fig:dataset}b and Appendix~\ref{sec:supp:dataset:hbv}).
For our experiments, we used a dataset with 128-bit vectors and a depth-6 tree.

To prevent the model from allocating a unique token to each branch (such as encoding generation 2 branches using `A, B, C, D' rather than `AA, AB, BA, BB'), we configured the model with 12 tokens grouped into 6 pairs, where only a single token from each pair may be added to a single sequence.
For example, a sequence may contain one or neither of tokens `A' and `B', but not both, resulting in a total of $6^3 = 216$ possible token sequences with a maximum length of 6.

\subsection{dSprites}
To evaluate the model with a richer modality, we used the dSprites dataset from \citet{higgins_beta-vae_2017}, which consists of images of shapes with varying features, including their size and position.
We adapted this dataset by adding color to the images, so that the shapes were assigned one of seven possible RGB color combinations with small variations, excluding black, which served as the background. 
We also modified the positions of the shapes so that they were distributed across a 4x4 grid with slight random adjustments to simulate real-world irregularities. 
We omitted the orientation attribute present in the original dataset, which can be difficult to infer in symmetrical shapes, focusing instead on the shape, color, size, and position features.
To ensure sufficient coverage in cases of overlapping token semantics, we equipped our model with 32 possible tokens and up to 7 tokens in each token sequence.

\subsection{Comparison models}

We used two popular models of emergent code through discrete bottlenecks for comparison: vector quantized variational autoencoders (VQ-VAE) \citep{van2017neural} and emergent communication models (EC) \citep{chaabouni2019anti, chaabouni2020compositionality, mu2021emergent}.
VQ-VAEs maintain one or more codebooks that contain a discrete set of vectors each associated with a token.
During the discretization step, a query vector is compared with all the vectors in the codebook and the nearest one is selected as the quantized form so that the resulting vector is the closest approximation of the query.
A typical implementation of VQ-VAE involves breaking an input image into patches and producing a discrete token for each individual patch.
However, because we were interested in descriptions that summarize the features of the whole image, we modified the model to form token sequences where all tokens are conditioned on the full image, thus allowing the model to form more holistic representations.

Emergent communication models  involve a speaker and a listener model where the speaker model is shown a target object with possible distractors and passes a discrete token sequence code to the listener model.
The listener model then uses the code to identify the target the speaker was shown among other distractors, thus encouraging the two models to learn a common language and for the speaker to be sufficiently descriptive for the listener.
The two models may be trained using reinforcement learning where both models are rewarded if the listener chooses the correct item \citep{chaabouni2019anti, chaabouni2020compositionality}, or using the Gumbel-Softmax to maintain differentiability using an approximate gradient \citep{mu2021emergent}.
We opted for the Gumbel-Softmax approach for training, conditioned the speaker on only the target item, and presented the listener with 49 distractor items.

In addition to the VQ-VAE and emergent communication models, we also include two hard-coded baselines for reference.
First, the fully-compositional (FC) baseline generates token encodings that include tokens that exactly correspond to the object features, \eg, `red' always maps to `A', so that the input features can be inferred from the generated sequences perfectly.
Second, the not-compositional (NC) baseline maps each input with shared features to a unique token sequence so that there is no semantic consistency between them.
These two models provide reference points for the highest and lowest possible compositionality given these datasets, assuming that there are no inputs with different features that map to the same token sequence.

\section{Results}

\subsection{Attractor dynamics}
We begin our analyses with the central feature of our model, the attractor dynamics, by confirming their contractive behavior near the sentence embeddings $\hat z_s$.
To do this, we perturb a state initially at $\hat z_s$ and unroll its trajectory to observe whether it converges back at the original attractor.
Starting with an initial condition $z_0$ of an input $x$, we generate a trajectory $z_0 \rightsquigarrow z_T$ and its corresponding token sequence $s$ evaluated at $z_T$.
Using the corresponding sentence embedding $\hat z_s$ as the reference attractor point, we add a perturbation $\nu$, characterized by a randomly oriented vector with a defined magnitude, to obtain a new initial state $z'_0 = \hat z_s + \nu$.
From $z'_0$, we simulate the dynamics again to produce a new trajectory $z'_0 \rightsquigarrow z'_{T'}$ over $T'$ time steps and subsequently measure its distance from $\hat z_s$.

\begin{figure}[t]
\centering
  \includegraphics[width=\linewidth]{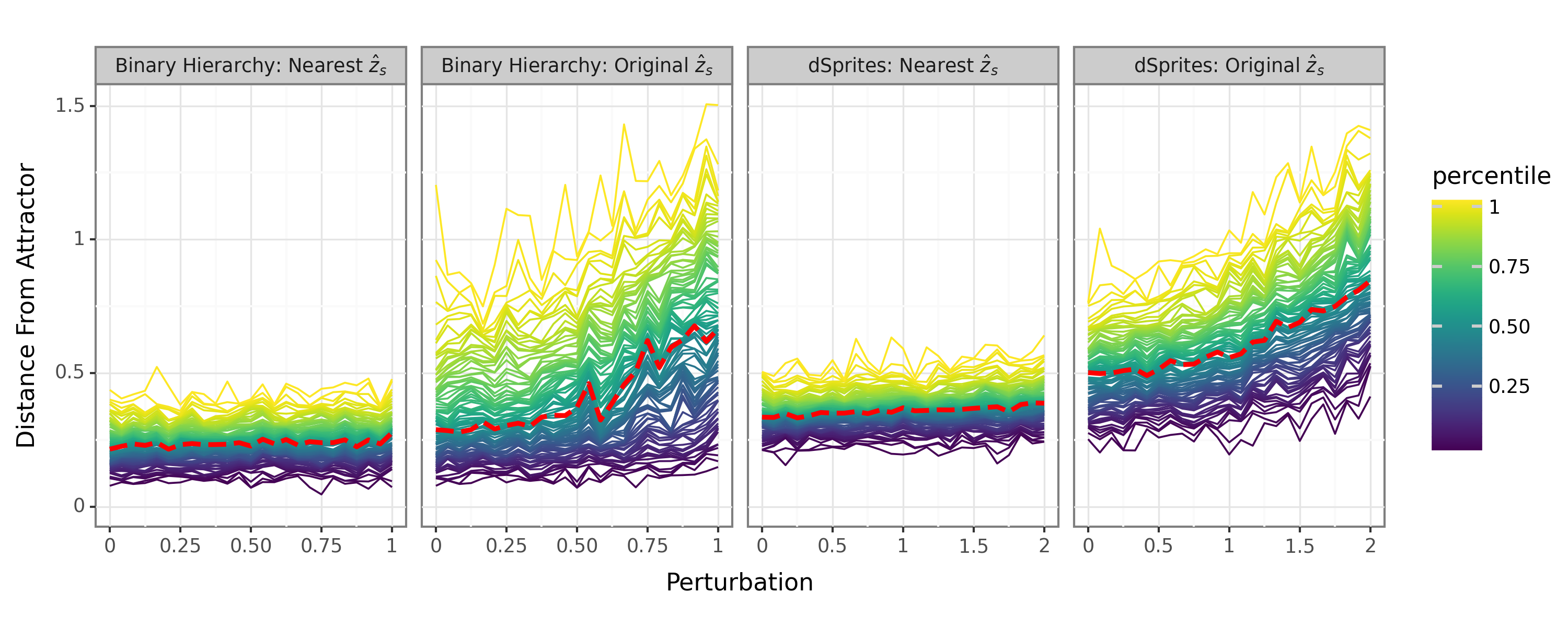}
\caption{
    Distance between a sentence embedding $\hat z$ and the post-dynamics state $z'_{T'}$ after adding varying degrees of perturbation.
    `Original $\hat z_s$' panels show distances to the original sentence embedding, while `Nearest $\hat z_n$' panels reflect distances to the closest sentence embedding post-dynamics.
    Thin lines represent distance percentiles and red dashed lines mark the median distances, with the inflection points suggesting the attractor basin boundaries.
    Higher percentile lines at small perturbations reflect trajectories that leave the attractor basin due to stochastic dynamics.
}
\label{fig:perturbation}
\end{figure}

Fig.~\ref{fig:perturbation} shows the distances from various degrees of perturbation.
Consistent with the properties of attractor points, the trajectories originating from perturbed states typically return to the vicinity of $\hat z_s$ when the perturbations are small.
While the inherent stochasticity of the system prevents $z'_{T'}$ from converging exactly on the sentence embedding $\hat z_s$, the median distance $\|z'_{T'} - \hat z_s\|$ remains relatively stable up to a critical perturbation size, which can be observed at an inflection point.
Furthermore, even with substantial perturbations, the distance $\|z'_{T'} - \hat z_n\|$ to the nearest sentence embedding $\hat z_n$ remains unchanged.
This observation suggests that even when a trajectory escapes the basin of the original attractor, it invariably settles within the basin of another nearby sentence embedding.

Fig.~\ref{fig:perturbation} also reveals an unexpected pattern: in the dSprites models, trajectories consistently terminate with a smaller distance to a different sentence embedding than to the original embedding $\hat z_s$, even when no perturbation is added.
We hypothesized that synonymous sentence embeddings could be clustering in a single attractor basin with minor local depressions at the location of each individual $\hat z_s$.
Such a structure could result in the discretizer selecting a sentence $s$ associated with an embedding $\hat z_s$ that, while semantically similar, is slightly further away.

\begin{figure}[t]
\centering
  \includegraphics[width=\linewidth]{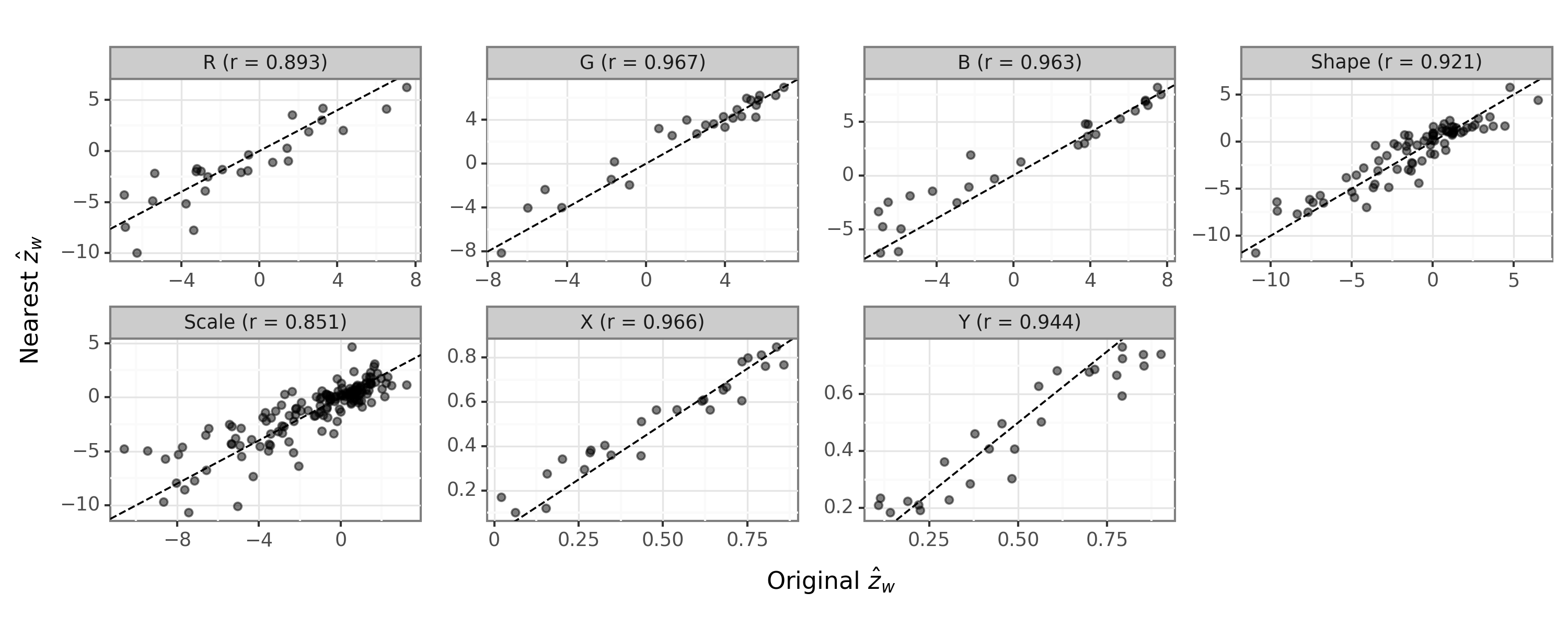}
\caption{
Correlation of predicted features between original and nearest post-trajectory sentence embeddings $\hat z_s$ and $\hat z_n$.
Each panel indicates a different image feature--R, G, B for color channels; shape; scale; X and Y for position--with the Pearson correlation coefficient (r). The dashed line represents the line of best fit.
}
\label{fig:nn_similarity}
\end{figure}

To test our hypothesis, we selected trajectories that started at the original attractor $\hat z_s$ without perturbation and terminated closer than the median distance from the starting point, but converged to a different embedding $\hat z_n$ so that $\hat z_n \neq \hat z_s$.
We then used a feature predictor trained on the input variables, \eg, color and shape, to decode the features from both $\hat z_s$ and $\hat z_n$.
As demonstrated in Fig.~\ref{fig:nn_similarity}, our findings indicate a strong correlation across all predicted features, substantiating our hypothesis.
The embeddings $\hat z_s$ and $\hat z_n$, despite being distinct, share a high degree of semantic similarity, collectively forming a mutual attractor basin that embodies a consistent concept.

\subsection{Emergent language}
\label{sec:language}

\subsubsection{Feature decodability}
\label{sec:language:decodability}
We evaluate the model's emergent language capabilities by training a decoder to predict feature labels associated with each input, utilizing the terminal point of sampled trajectories $z_T$ along with its corresponding sentence embedding $\hat z_s$.
In the HBV dataset, these labels trace the hierarchical path from the root node to the prototype node of the input vector.
We evaluated the decoder performance on the first four branches of the path where the class features retain sufficient signal over the noise.
For the dSprites dataset, we report the accuracy measures on object color and shape features, which we treat as categorical features, and the root mean squared error (RMSE) measures for the size and position features, which we treat as continuous features.

To examine the model's ability to generalize beyond the training distribution (within-distribution; WD), we systematically withheld subsets of the dataset (see Fig.~\ref{fig:dataset}) for out-of-distribution (OOD) evaluation.
In the HBV dataset, we excluded one prototype per generation from the third generation onward, excluding those with an ancestor already held out. 
In the dSprites dataset, we withheld all images featuring three specific combinations of features. 
The `position' set encompassed all images with shapes positioned within two different quartiles of the X and Y values, covering 1/16th of the dataset within a square region. 
The `position + color' set included images with shapes in another square region, but exclusively featuring yellow shapes. 
Finally, the `position + shape' set comprised of all images with magenta ovals.

As shown in Fig.~\ref{fig:compositionality} (red bars), the attractors model and both of the comparison models outperform  the NC baseline (higher accuracy, lower MSE), though none generalize perfectly (FC baseline).
While the three models are comparable on most accounts, we observe a substantial improvement on out-of-distribution generalization by the attractors model in the dSprites dataset, reaching almost full generalization on withheld shapes.



\subsubsection{Representational similarity analysis}
Representational similarity analysis (RSA), also referred to as topographic similarity (TopSim), is a popular metric for evaluating compositionality in emergent languages.
RSA compares how relationships and shared attributes across two distinct representational systems are encoded by computing a pairwise-distance matrix of a common set of inputs and using the Spearman correlation between the matrices as the similarity score.
In our analysis, the distance between two sequences of tokens (represented as sets of tokens in our study) is quantified by converting them into binary vectors and counting the number of matching bits. 
For the dSprites images, distance is assessed by counting the categorical features (shape and color) that differ and calculating the absolute distance for continuous features (size, x-position, y-position).
Applied to the dSprites dataset, we recorded RSA scores of 0.525 for our attractors model, 0.156 for the emergent communication model, and 0.314 for the VQ-VAE model respectively.

However, despite RSA's prevalence, it presupposes linear relationships between representations, limiting its effectiveness in analyzing complex linguistic structures. 
For example, consider the base-2 representations of the numbers 0, 7, and 8, which are \texttt{0000}, \texttt{0111}, and \texttt{1000}, respectively. 
A comparison of these base-2 representations as token sequences would misleadingly suggest that numbers 0 and 8, sharing three bits, are more similar than numbers 7 and 8, which share no bits. 
Thus, RSA and other simple distance-based metrics may fail to capture the nuances of hierarchical relations, which is necessary for the HBV dataset.

\subsubsection{Feature-wise composition}

To avoid the linearity assumption of RSA, we measure compositionality by asking how well we can describe an object one feature at a time by constructing a sentence one token at a time, a process which we refer to as feature-wise composition (FWC).
For instance, if we want to describe a red object, we look for the token that, according to a trained feature decoder, maximizes the likelihood of inferring `red' when added to the token sequence.
For the HBV, we consider the top-level feature $y_1$, \ie, the first branch taken at the top level of the binary tree.
Then, using a trained decoder, we find the token $w_1$ that maximizes the likelihood $P(y_1 \mid w_1)$ using a trained feature predictor.
We repeat this process going down the tree, at each step selecting the token that maximizes the joint likelihood of all previously considered features: $P(y_1, \ldots, y_i \mid w_1, \ldots, w_i)$.
We apply a similar strategy to the dSprites dataset, but randomly choose the order of seven features considered since there are no hierarchical dependencies.
Because a dSprite image contains both categorical and continuous features, we select tokens that minimize the overall prediction loss at each step.

We assess the compositionality by using the token sequences to predict the input features: the first four class levels for the HBV dataset and the image attributes for dSprites.
To eliminate bias that arises from using the same decoder to both produce and evaluate the generated sentences, we use two separately trained feature decoders.
We then compare the prediction performance of sentences composed using FWC and sentences sampled directly from the model.
A higher model score than the FWC score could indicate non-compositional encodings; however, it is possible that the language is compositional in a different way.
Conversely, a lower model score would suggest that the model's sampling policy does not necessarily always reflect the production rule.

\begin{figure}[t]
\centering
  \includegraphics[width=\linewidth]{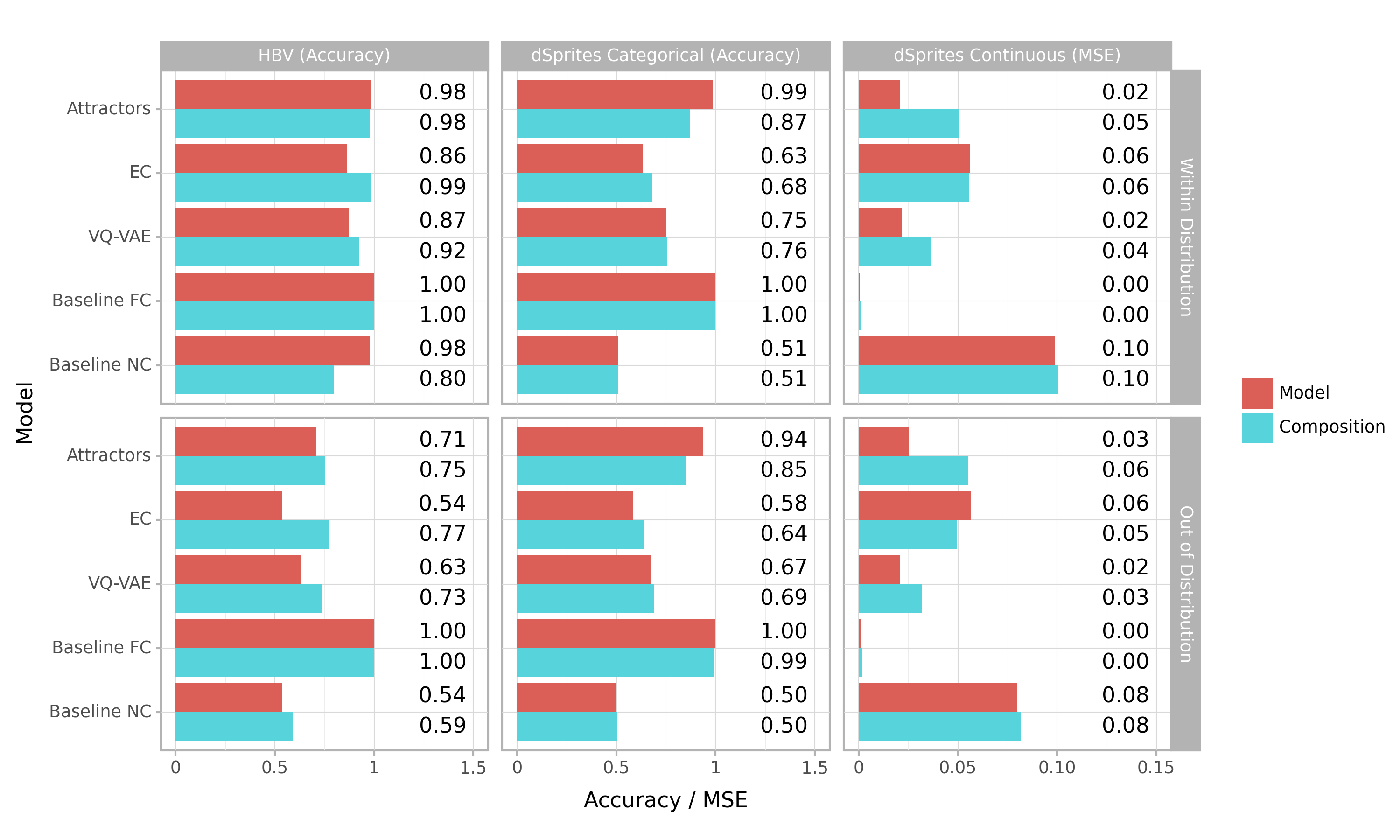}
\caption{
Feature predictor performance.
Left and center panels show accuracy scores for categorical features (higher is better). Right panels show MSE for continuous features (lower is better).
Top panels show scores for within-distribution sets and bottom panels show scores for out-of-distribution sets.
Red bars indicate scores for features decoded from token sequences drawn directly from the model, whereas blue bars indicate scores for sequences constructed using feature-wise composition.
The lower accuracy scores of the EC and VQ-VAE models than the NC baseline in the HBV dataset is a consequence of multiple inputs mapping to identical token sequences, an error not accounted for in the NC baseline.}
\label{fig:compositionality}
\end{figure}

As shown in Fig.~\ref{fig:compositionality}, we observe small or no differences in feature prediction accuracy between the sentences sampled from the model and sentences generated using FWC for within-distribution inputs.
Similarly high accuracy and low MSE scores in both the model-generated and FWC sequences suggest that accurate descriptions generated by the model are consistent with a highly compositional production rule.
As expected, the FWC score also drops in the out-of-distribution set, though interestingly, sometimes not as much as the model score.
This may be a consequence of the rigid production rule that constrains, and therefore regularizes, the FWC generative process.

\subsection{Sample diversity}
\label{sec:diversity}
A fundamental attribute of PLoT models is their capability to learn entire distributions rather than singular point estimates, thereby capturing uncertainty and diversity. 
These distributions, grounded in Bayesian statistics, provide rational accounts of human cognition, corroborated by experimental data \citep{goodman2008rational, kemp2008discovery, goodman2011learning, piantadosi2012bootstrapping, goodman2016pragmatic}. 
In our study, we assess the diversity of sample trajectories generated by our model and evaluate how effectively the distributions of token sequences encapsulate the informational content of the external inputs they represent.

We begin our analysis of distributional diversity by measuring the entropy of feature predictions decoded from the latent representations with three important points.
First, we measure the conditional entropy linked to individual inputs $H(\cdot \mid x)$—analyzing the range of descriptions our model generates for a particular item, rather than the overall entropy of the samples $H(\cdot)$, which would also reflect the underlying entropy of the data distribution. 
This helps us evaluate the model’s capacity to generate diverse yet plausible outputs from a given input.

Second, we focus on the entropy of semantic content rather than simply the diversity of sentence generation, which could be artificially inflated by expanding the model's vocabulary. 
For example, in response to an image of a red square, our goal is to assess the model’s ability to differentiate and generate distinct concepts, such as `red', `square', and `red square'.
Therefore, we ignore semantically equivalent variations like `red square' and `scarlet square', thus favoring entropy of encoded object features $P(\text{feature} \mid x)$ over the token sequences $P(w \mid x)$.

Lastly, to measure concept diversity, we distinguish between models that consistently output parameters representing a single concept and those that generate a variety of concepts. 
To illustrate what we mean, consider two simple probabilistic models that output parameters for a Beta distribution.
One model always outputs $\{ \alpha = 1, \beta = 1 \}$, \ie, a uniform distribution from 0 to 1. Although the resulting distribution itself has maximal entropy, we would view this model as generating a single, homogeneous concept—indicative of minimal conceptual diversity. 
In contrast, a model that generates $\{ \alpha = 1000, \beta = 1 \}$ and $\{ \alpha = 1, \beta = 1000 \}$ with equal probability effectively samples from two distinct concepts, so that although the two resulting Beta distributions have extremely low entropy, this has a higher concept diversity. 
Applying this approach to our analysis, we measure the entropy directly on the distribution of outputs from the feature predictor model $P(p_{\rm feature} \mid x)$.

We estimate the empirical differential entropy of the feature decoder outputs using the Kozachenko-Leonenko estimator \citep{kozachenko1987sample}. 
This method approximates entropy by computing nearest-neighbor distances to assess local data concentration, thereby providing an estimate without any assumptions about the underlying data distribution.
To aid in interpreting these values, we also report the standard deviation (SD) $\sigma$ of an isotropic Gaussian with the same dimensionality and entropy for comparison.
In our analysis of the HBV dataset, we obtain a mean differential entropy of $-18.198$ with 95\% of scores between $[-34.199, -5.203]$ nats, corresponding to a standard deviation of $\sigma = 0.088$ $[0.36, 0.181]$.
This result is significantly higher than the $-173.66$ $[-218.62, -62.666]$ nats ($\sigma = 0.000$ $[0.000, 0.007]$) observed in the EC model.
The VQ-VAE, which only provides deterministic encodings, has an undefined ($-\infty$) differential entropy.
For the dSprites dataset, the mean entropy is $-9.415$ $[-16.231, -3.832]$ nats ($\sigma = 0.085$ $[0.040, 0.158]$), again significantly higher than the $-49.973$ $[-109.607, -11.785]$ nats ($\sigma = 0.001$ $[0.000, 0.065]$) of the EC model.

This analysis provides strong evidence of sample diversity. 
However, assessing how well the distribution of sampled token sequences matches the information posterior remains a complex challenge. 
The emergent nature of the token sequences complicates establishing a normative distribution without a translational framework. 
Although it is feasible to define a normative distribution in feature space and use feature decoders to map the token sequences, this approach does not necessarily ensure normativity in latent space.
This is particularly problematic if multiple distinct regions represent the same concept -- a likely scenario given the combinatorially large number of possible sentences. 

Despite these limitations, we find that the simplicity of the HBV dataset and the relatively small number of possible sentences in the HBV models allow us to evaluate the goodness-of-fit, even without a theoretic guarantee.
Recall the model's objective: sample trajectories starting from a maximally informative initial state $z_0$ of a high dimensional input $x$, such that the sampling is proportional to the mutual information between $x$ and the terminal state $z_T$.
We begin by estimating the upper bound of representative capacity of $x$ in latent space, measured by the number of correctly predicted bits from $z_0$ by a trained reconstruction model, denoted as $c_0$.
This quantity is then compared to the number of bits correctly predicted from $z_T$ (denoted as $c_T$), so that the difference $d = c_0 - c_T$ represents the amount of information lost during discretization.
Each correctly predicted bit corresponds to a twofold reduction in entropy, allowing us to hypothesize a normative pattern in the probability distribution of information loss.
Specifically, we expect the probability of $d = 1$ to be half that of $d = 0$, and the probability of $d = 2$ half of that, etc., thereby forming a geometric distribution with a probability parameter of $0.5$.
We test this hypothesis by comparing the empirical distribution of $d$ from our model's outputs against the theoretical distribution: ${\rm Geometric}(0.5)$.

\begin{figure}[t]
  \centering
  \includegraphics[width=\linewidth]{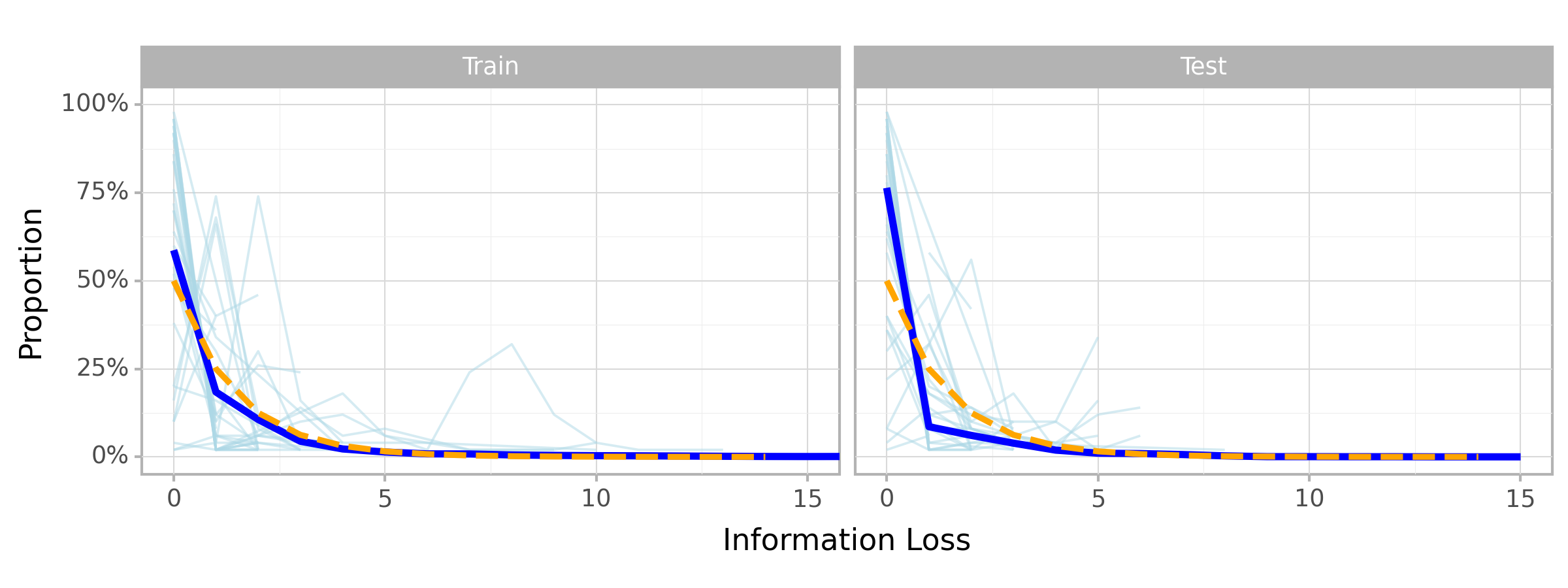}
  \caption{Distribution of the amount of information lost during discretization, measured as the difference in number of input bits correctly predicted when decoded from $z_0$ and $z_T$. Light blue lines indicate sample distributions for individual inputs. Dark blue lines indicate the marginal distribution across inputs.
  Dashed orange lines indicate the theoretic distribution $Geometric(0.5)$.}
\label{fig:hbv_distribution}
\end{figure}

As illustrated in Fig.~\ref{fig:hbv_distribution}, the empirical distribution derived from the training samples closely aligns with the predicted distribution.
The empirical parameter $p$ for the within-distribution set is $0.480$, which is very close to the hypothesized $0.50$, and the KL-divergence from ${\rm Geometric}(0.5)$ is $0.041$.
Although the overall shape of the distribution is consistent for the out-of-distribution set, the model more strongly favors samples with lower information loss, resulting in $p = 0.621$ with a KL-divergence of $0.183$.

\section{Discussion}

We have presented a novel stochastic dynamical systems model of discrete, symbolic thought that provides a sub-symbolic account.
Similar to other emergent language models \citep{van2017neural,chaabouni2019anti,chaabouni2020compositionality,mu2021emergent}, our model produces discrete token sequence representations of the data without any domain specific knowledge or pre-specified primitive concepts.
This language reflects not only the semantics and features that characterize the dataset, but utilizes compositionality to recombine simpler concepts into more complex ones.
Using a hierarchically generated dataset of binary vectors and the colorized dSprites dataset containing images of simple geometric shapes, we have shown that our model demonstrates competitive performance against established models like the VQ-VAE \citep{van2017neural} and emergent communication \citep{mu2021emergent} models.

What distinguishes our model from these predecessors, however, is the method of discretization via stochastic attractor dynamics.
Unlike other discrete-bottleneck models that select an entry from a lookup table of vectors, our model forms discrete representations through the system dynamics, starting from information-rich states and gradually settling on attractor regions over time.
Thus, our model uniquely offers a unified framework for both symbolic and sub-symbolic modes of thought from stimulus onset to its stable encoding.
This aligns with observed brain mechanisms in various cognitive functions \citep{khona2022attractor}, such as short-term memory \citep{inagaki2019discrete}, head orientation \citep{chaudhuri2019intrinsic}, and pattern separation \citep{lin2014sparse}.

Moreover, our model learns to generate diverse sets of discrete sequences for any given input, where the resulting distribution reflects the Bayesian posterior that reflects the mutual information between the information-rich inputs and their discretized forms.
Although this property is characteristic of continuous latent samplers such as the VAE \citep{Kingma2014AutoEncodingVB} and diffusion models \citep{sohl2015deep}, it is generally not guaranteed in discrete latent samplers.
By overcoming this limitation, our model aligns with the Bayesian tradition of cognitive modeling \citep{goodman2008rational, kemp2008discovery, goodman2011learning, goodman2016pragmatic}, \ie, the probabilistic language of thought (PLoT) \citep{goodman2015concepts}, which learns distributions over program-like symbolic compositions according to a Bayesian posterior. 
The dual alignment to both continuous neural mechanisms and PLoT helps bridge a long standing gap between connectionist (sub-symbolic) \citep{mcclelland2010letting} and Bayesian (symbolic) \citep{griffiths2010probabilistic} models of cognition.

Unifying these two approaches introduces new avenues to explore computational models of cognitive phenomena that have previously been difficult to capture using existing models.
First, like other neural networks with discrete bottlenecks, our model eliminates the need to pre-define concepts and composition rules, instead allowing these to emerge during the learning process.
However, our model is uniquely able to learn using Bayesian principles so that the resulting distributions are not arbitrary or purely incidental.
Second, by using a dynamical system where the initial information-rich state gradually settles to its lossy discrete representation over time, the model reveals the intermediate trajectory between these two endpoints.
This transparency is crucial for understanding the mechanisms underlying lossy representation in cognitive systems that continuously, albeit rapidly, unfold over time, such as schema-based memory \citep{brewer1981role}, categorical perception \citep{liberman1957discrimination, goldstone2010categorical, mcmurray2002gradient}, and attention \citep{corbetta2002control}.
Lastly, a neural dynamical systems model may offer insights into similar attractor-based mechanisms in the brain \citep{khona2022attractor}, such as in decision making \citep{inagaki2019discrete}.


Despite the advantages, the model is not without its limitations. 
Although our model aims to reduce the gap between discrete symbolic and sub-symbolic processes, it presupposes the existence of symbols \textit{during training} with a decoding mechanism that translates latent states to the symbolic sequences.
One interpretation for this mechanism is an output system that emerges as a consequence of an environment that emphasizes distinctive features, such as phonological primitives that are inherently distinct from one another to enhance communicative clarity \citep{jakobson1956phonology}.
While it may be possible to describe this as a prerequisite cognitive construct, a more complete account of emergent symbolic representations would benefit from a learning mechanism that does not rely on any explicit discretization step, but instead learns it end-to-end.
Future research could explore methods for symbol generation that do not involve explicit discretization into tokens, even during training.

In conclusion, we propose a general framework for modeling the organization of symbolic representations through sub-symbolic mechanisms.
The methods and results presented in this paper demonstrate the model's efficacy in simple domains, opening up future research applying the model to more sophisticated tasks and datasets.
It may also be interesting to investigate what the system dynamics can reveal about how cognitive systems organize information gradually over time rather than through a single feed-forward step.
Our approach offers a unique opportunity to understand how neural trajectories constrain the system and how the dynamics evolve during training to accommodate a temporal process.
These avenues may yield insights that take us a step closer towards understanding the algorithms and neural mechanisms that underlie human symbolic cognition.

\clearpage
\bibliographystyle{abbrvnat}  
\bibliography{neurips_2023}  

\clearpage
\appendix

\section{Glossary}

\subsection{Symbols}
\begin{itemize}
    \item $F$: the \textit{flow} value associated with a state $z$
    \item $L$: the length of a token sequence
    \item $R$: the reward function that represents the unnormalized target density for the GFN; this quantity may be expressed as an energy function using the relationship $R = e^{-\mathcal{E}}$
    \item $T$: the length of a trajectory
    \item $Z$: the normalizing constant used to scale the reward to a probability
    \item $g$: a flow correction term used in the forward-looking GFN objective
    \item $s$: a discrete token sequence composed of word tokens $w$
    \item $t$: a time step in a trajectory
    \item $w$: a \textit{word} or token in the token sequence $s$
    \item $x$: an exogenous input, such as a binary vector in the Hierarchical Binary Vectors dataset or an image in the dSprites dataset
    \item $y$: a feature vector, such as the color of the shape in the dSprites dataset
    \item $z$: a state in the latent space
    \item $z_0$: the embedding vector of an input $x$ that shares maximal mutual information; this is also the initial condition (IC) of a trajectory and the representation used in modules that reference $x$, \eg, the reward function $R_\phi(s, x)$, which can be read as $R_\phi(s, z_0)$
    \item $z_T$: the terminal state of a trajectory $z_0 \rightsquigarrow z_T$
    \item $\hat z_s$: the emebedding vector of a token sequence $s$
    \item $\alpha$: the probability of sampling off-policy during exploration
    \item $\theta$: the set of parameters trained during the E-step; also the set of parameters trained using GFN objectives
    \item $\phi$: the set of parameters trained during the M-step
\end{itemize}

\section{Training details}
\label{sec:training_details}
Here, we include some of the mathematical details and computational methods involved in training the model that were not included in the main text.

\subsection{E-step}
In addition to the modules described in S\ref{sec:training}, an essential component of the model is the `flow' module, briefly mentioned in the main text, which measures the expected future reward from a state $z_t$, denoted as $F_\theta(z_t, t, x)$.
Using this quantity, the relationship between two states $z_t$ and $z_{t+1}$ may be related through the transition functions that form a detailed balance constraint \citep{bengio2021foundations}:
\begin{equation}
\label{eq:attractors:detailed_balance}
    F_\theta(z_t, t, x) P_\theta(z_{t+1} \mid z_t, t, x) = F_\theta(z_{t+1}, t+1, x) P^\leftarrow_\theta(z_t \mid z_{t+1}, t+1, x).
\end{equation}

This quantity is closely tied to the reward function, which may be used as an inductive bias by defining a `forward-looking' flow $g_\theta(z,x,t)$ that applies a correction to the reward quantity \citep{pan2023better}, constrained to equal 0 when $t=T$. 
\begin{equation}
    F_\theta(z, t, x) = \sum\limits_s P(s \mid z, t, x) R_\phi(s, x) = R_\phi(s, x) \dfrac{P_\theta(z \mid \hat z_s, x)}{P_\theta(s \mid z, x)} g_\theta(z, t, x).
\end{equation}
The $s$-independent reward $R(z, x)$ denotes the expected reward of $\mathbb{E}_s [R(z, s, x)]$ over symbol sequences $s \in \mathcal{S}$, which may be estimated by sampling $s \sim P_\theta(s \mid z)$ and evaluating
\begin{equation}
    R(z, x) 
    = R_\phi(w, x) P^\leftarrow_\theta(z \mid \hat z_s, T+1, x) \prod\limits_{i=0}^{L-1} \dfrac{P^\leftarrow_\theta(s_i \mid s_{:i+1}, z)}{P_\theta(s_{i+1} \mid s_{:i}, z)},
\end{equation}
where $L$ is the length of the token sequence.

Putting it all together, we train the model by balancing the two sides of Equation~\ref{eq:attractors:detailed_balance} using mean squared error between their log-quantities, which yields the loss function of a given trajectory $z_0\rightsquigarrow z_T$:
\begin{equation}
\label{eq:traj_loss}
    \mathcal{L}_{\rm E} = \sum\limits_{t = 0}^{T-1} \left ( \log \left ( 
    \dfrac{R(z_t, x)}{R(z_{t+1}, x)} 
    \dfrac{g_\theta(z_t, x, t)}{g_\theta(z_{t+1}, x, t+1)} 
    \dfrac{P_\theta(z_{t+1} | z_t)}{P_\theta^\leftarrow(z_t | z_{t+1}, x, t+1)} \right ) \right )^2.
\end{equation}

Note that the forward dynamics in Eq.~\ref{eq:traj_loss} are input- and time-independent ($x$ and $t$), which reflects the model used in the main text, whereas the detailed balance constraint in Eq.~\ref{eq:attractors:detailed_balance} are not.
While we introduced these modifications for better cognitive plausibility, mathematically, they may be viewed as model regularizations.

\paragraph{M-step} 
As described in the main text, the optimization during the M-step involves optimizing $P_\phi(z_0 \mid x)$ using a variational autoencoder \citep{Kingma2014AutoEncodingVB} with an additional KL-divergence term to increase the mutual information between $x$ and $s$.

\begin{equation}
\label{eq:mstep_loss}
    \mathcal{L}_{\rm M} = 
    \underbrace{\text{recon}(x, z_0) + D_{\rm KL}( P_\phi(z_0 \mid x) \| P(z_0))}_{\text{VAE}} 
    + \underbrace{\widehat{D_{\rm KL}}( P_\phi(z_0 \mid x) \| P_\phi(z_0 \mid \hat z_s))}_{\text{mutual information}}.
\end{equation}


We also found it effective to apply weighted gradient updates, similar to the VQ-VAE objective, that favor modifying $z_0$ over $z_s$ using a stop-gradient (SG) with $\beta = 0.25$:
\begin{equation}
    \widehat{D_{\rm KL}}( P_\phi(z_0 \mid x) \| P(z_0 \mid \hat z_s)) = D_{\rm KL}( P_\phi(z_0 \mid x) \| P^{SG}_\phi(z_0 \mid \hat z_s)) + \beta D_{\rm KL}( P^{SG}_\phi(z_0 \mid x) \| P_\phi(z_0 \mid \hat z_s)). 
\end{equation}
The decodability term $P_\phi(s \mid z_T)$ in Eq.~\ref{eq:reward} uses the relation $P(s \mid z_T) \propto P(z_T \mid s) \cdot P(s)$, where $P(z_T \mid s)$ is computed using $P_\phi(z_0 \mid \hat z_s)$.

\subsection{Exploration}
GFlowNets, much like RL models, require an exploration policy to encourage discovery of new modes.
However, the constraints of our model places unique challenges in off-policy exploration.
Here, we describe several challenges and possible approaches to address them.

\paragraph{Off-policy sampling} We allow off-policy sampling for the dynamics model by sampling with some small probability $\alpha_{dyn}$ a random next state near the current state uniformly within the bounded step region.
We apply similar off-policy sampling for the discretizer with a small probability $\alpha_{disc}$ of sampling uniformly among all available tokens at each token-step.

Unfortunately, this approach alone, while simple and effective in generating diverse samples, can introduce extreme variability that can significantly obstruct optimization due to how the reward function is used.
Because our reward estimate $R(z, x)$ relies on using sampled $w$ as a reference point, its estimate can be highly variable depending on the $w$ considered depending on the distance between $z$ and $\hat z_s$, where $P^\leftarrow_\theta(z \mid \hat z_s, x)$ can be several orders of magnitude larger for a distant $\hat z_s$ than a nearby one.
While this in theory should be mitigated by the weighting term $P(s \mid z, x)$, maintaining an accurate estimate is intractable in practice when the space of $\mathcal{S}$ is exponentially large and the probability of sampling any $S$ off-policy is therefore exponentially small.

Our solution is to decouple the exploration for the discretizer from the dynamics model and sample $s \sim P(s \mid z, x)$ on-policy when computing the reward, since on-policy estimates are more likely to be accurate.
The discretizer is then trained separately using the trajectory balance objective in Eq.~\ref{eq:discretizer_loss} with off-policy sampling (sampling $s_i$ uniformly with probability $\alpha$) with an additional auxiliary model that estimates the partition function $Z_\theta(z, x)$.
This allows the loss variance during exploration to be absorbed by the discretizer model rather than the dynamics model, significantly improving training stability.
\begin{equation}
\label{eq:discretizer_loss}
    \mathcal{L}_{\text{disc}} = \left ( \log \dfrac{Z_\theta (z, x) P_\theta(s_1 \mid z, x) \dotsc P_\theta(s_L \mid s_{:L-1}, z, x)}{R_\phi(s, x) P_\theta(z \mid \hat z_s, x) P(s_{L-1} \mid s, z, x) \dotsc P(s_1 \mid s_{:2}, z, x)} \right )^2
\end{equation}

\paragraph{Flow correction} As the model learns to form attractor basins, the basins form steep local minima that trap sample trajectories to basins with low $R(s, x)$.
Moreover, because the ``flow'' $F(z, x, t)$ is learned implicitly by applying a correction on the energy function, this further biases the exploration towards the nearest attractors rather than attractors that reflect the input.
We help mitigate some of the difficulty by drawing on the fact that corrections are larger for earlier states in the trajectory when it has yet to reach an attractor basin, and apply a simple linear scaling factor to $g(z, x, t)$ based on the number of steps remaining in the trajectory, $T - t$.

\paragraph{Weighted replay buffer}
Because the reward model is trained using samples from the GFN, and the GFN is trained to sample according to the reward model, a token that has been out of circulation is unlikely to re-emerge later in training.
Consequently, the model is liable to collapse to a local minimum that only uses a small subset of tokens in the vocabulary that it was equipped with.
We address this by maintaining a replay buffer of trajectories and corresponding token sequences. 
Each entry in the buffer is then weighted by the inverse of how frequently the particular token sequence appears in the buffer and how often each token in the sequence appears in the buffer.

\paragraph{Wake-sleep} 
We take advantage of the fact that all trajectories are evaluated against discrete objects $s$, which allows us to apply the wake-sleep algorithm described in \citep{hu2023gfnem}.
The wake-sleep algorithm uses the backward policy to generate trajectories for modes that the forward policy may have missed by starting at terminal states and ending at starting states.
Given an initial condition $z_0$ and a target token sequence $s$, we sample a trajectory from its corresponding $\hat z_s$ using the backward trajectory $P^\leftarrow_\theta(z_{t-1} \mid z_t, z_0, t)$.
Although the backward policy is conditioned on $z_0$ and is intended to terminate at $z_0$, the model may produce a trajectory that terminates elsewhere at $z^\leftarrow_0 \neq z_0$ during training.
We therefore we correct the trajectory by shifting it such that its endpoints are $z_0$ and $\hat z_s$, resulting in an inductive bias proportional to the distance between the sampled endpoint $z^\leftarrow_0$ and the intended endpoint $z_0$.

We sample $z_0$ and $s$ pairs in two ways.
One, we sample $s$ from a prior $P(s)$ and a $z_0$ from $P_\phi(z_0 \mid \hat z_s)$ learned during the M-step.
While this has the advantage of producing maximal diversity, it has the limitation of generating samples that are outside the data distribution, either due to the regions in the latent space that they occupy, or due to the semantic and syntactic structure of $\mathcal{S}$ that might make certain token combinations less plausible, such as a square circle.
Two, we use the token sequences produced through forward trajectories with off-policy sampling described above.
Given a trajectory with $T$ steps, we draw a token sequence $s_t \sim P_\theta(s_t \mid z_t)$ at each point of the trajectory and sample a single sequence $s$, weighted according to their reward values $R_\phi(s, x)$.

\paragraph{Energy gradient}
One set of exploration methods that has been shown to be effective in GFN diffusion models \citep{sendera2024diffusion} involves the use of the energy gradient. 
Langevin parameterization uses a gradient-informed policy \citep{zhang2021path} that allows the model to follow the gradient of the energy function as the starting basis of the policy that the model gradually learns to temper in order to learn a multi-modal posterior.
The Metropolis-adjusted Langevin algorithm, which applies a similar strategy to the Metropolis-Hastings algorithm for MCMC by using an acceptance rule based on the energy function.

In practice, we have found gradient-based methods to be unhelpful and potentially harmful for our model.
Recall that our reward function $R_\phi(z, x)$ involves sampling a $w$ to compute a single-sample weighted estimate.
However, given $w$, the only reward function reduces to a distance measure between the latent state $z$ and the sentence embedding $\hat z_w$.
Thus, while gradient-based methods can be used with respect to the sampled $w$, they would do no better than using a direct line to $\hat z_s$ as the inductive bias, which can interfere with learning more interesting dynamics.
Two possible alternatives are to use multiple $w$'s to estimate $R(z, x)$ or to train a model to estimate $R(z, x)$ directly, though the latter approach has not been effective in our experiments.

\section{Dataset details}

\subsection{Hierarchical binary vectors (HBV)}
\label{sec:supp:dataset:hbv}

\begin{figure}[t]
\centering
\includegraphics[width=\linewidth]{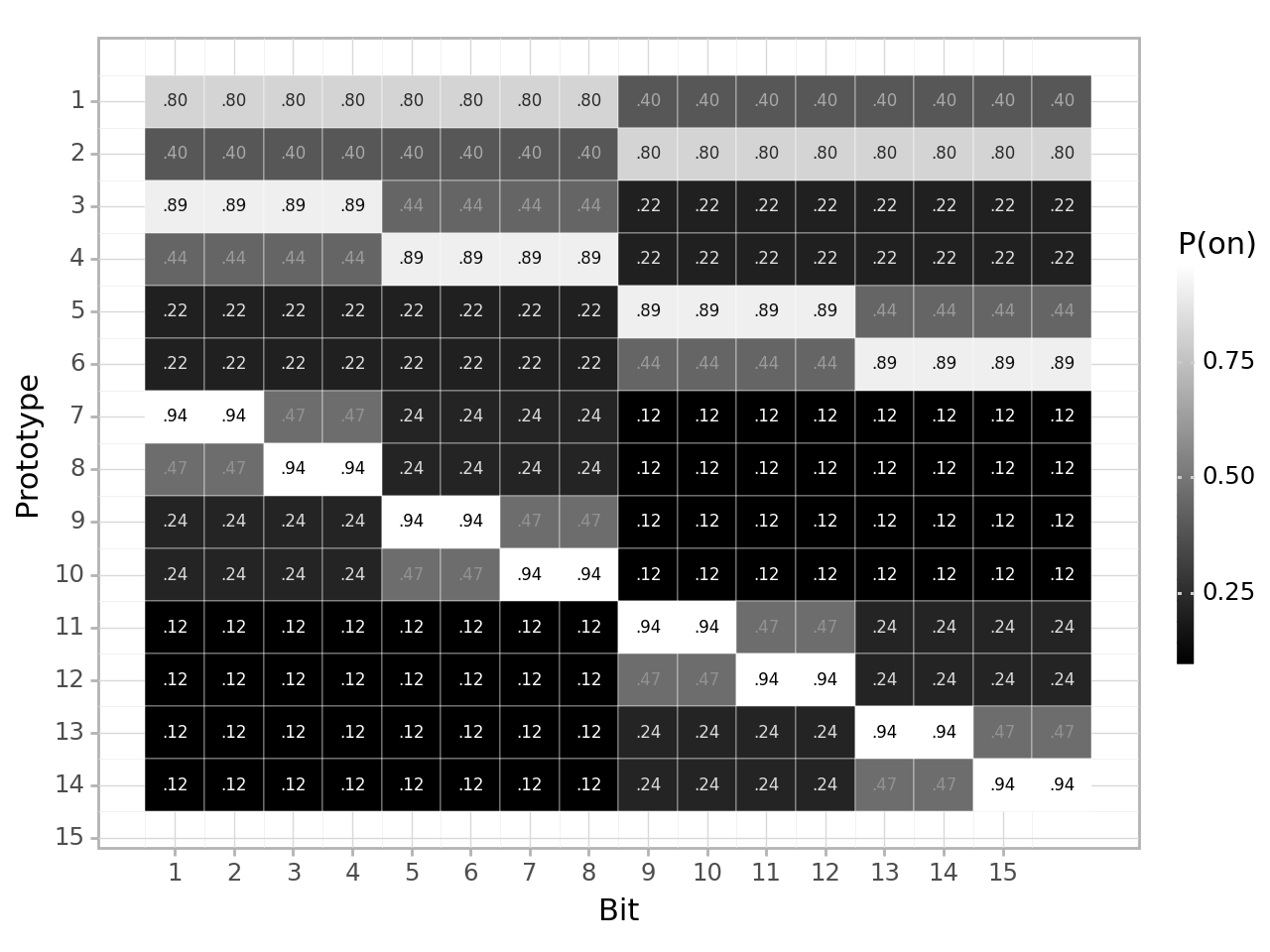}
\caption{
The probability of sampling 1 for each bit in a $D=3, L=2$ HBV dataset.
}
\label{fig:supp:bhv_probs}
\end{figure}

To sample exemplars from a prototype, we independently sample the value of each bit using the following formula.
For a prototype bit sequence at depth \(d'\), let \(d_k\) represent the depth of the most recent ancestor where the value of the \(k^{th}\) bit of its prototypical sequence is a \texttt{1}. 
The probability of the \(k^{th}\) bit in the sequence sampling a \texttt{1} is calculated as \(P(\text{bit}_k = 1) = \frac{2^{1+d_k}}{1 + 2^{1+d'}}\).
As shown in Fig.~\ref{fig:supp:bhv_probs}, the probability matrix reflects a noisy rendition of the prototype vectors shown in Fig.~\ref{fig:dataset}a that reflects a 1:2 ratio for cross-category bits.

\subsection{dSprites}
The dSprites dataset  consists of images containing a single square, heart, or ellipse.
Although the dataset as used in the original paper contains white shapes of 6 possible linearly spaced sizes, 40 orientations, and $32 \times 32$ $(x, y)$ positions in a grid, we introduced small modifications to explore the relationship between input modes and attractor dynamics.
The statistics of the position variables were modified such that the frequency of $x$ and $y$ values follow a [1, 2, 3, 4, 5, 4, 3, 2, 1] pattern across 31 positions (one is removed) with 4 evenly spaced modes, effectively producing a $4 \times 4$ grid with jittered positions.
We also added color to the dataset, where the modes are the 7 combinations of binary RGB values excluding black (0, 0, 0), and injected noise by sampling random values from $HalfNormal(\sigma^2 = 0.04)$.

\section{Pseudocode}
\label{sec:supp:attrators:pseudocode}

\begin{algorithm}
\caption{Training Loop}\label{alg:attractors:training_loop}
\begin{algorithmic}[1]
\Repeat \Comment E-step
    \State Populate replay buffer \Comment Alg.~\ref{alg:attractors:replay_buffer}
    \Repeat 
        \State E-step inner loop: dynamics \Comment Alg.~\ref{alg:attractors:e_step_dynamics}
        \State E-step inner loop: discretizer \Comment Alg.~\ref{alg:attractors:e_step_discretizer}
    \Until stopping condition
\Until stopping condition
\Statex
\Repeat \Comment M-step
    \State Populate replay buffer \Comment Alg.~\ref{alg:attractors:replay_buffer}
    \Repeat 
        \State M-step \Comment Alg.~\ref{alg:attractors:m_step}
    \Until stopping condition
\Until stopping condition
\end{algorithmic}
\end{algorithm}

\begin{algorithm}
\caption{Populate replay buffer}\label{alg:attractors:replay_buffer}
\begin{algorithmic}[1]
\Require $x \in \mathcal{X}$
\Statex
\State Sample $z_0 \sim P_\phi(z_0 \mid x)$ \Comment{Sample trajectory}
\Repeat 
    \State Sample $z_{t+1} \sim P_\theta(z_{t+1} \mid z_t)$
\Until $t >= T$ or convergence condition
\Statex
\State Sample $s_0 \sim P_\theta(s_0 \mid z_T, x)$ \Comment{Sample token sequence}
\Repeat 
    \State Sample $s_{l+1} \sim P_\theta(s_{l+1} \mid s_0, \ldots, s_{l}, z_T, x)$
\Until end-of-sequence token $s_L$
\State $s \leftarrow s_0, \ldots, s_L$
\Statex
\If{E-step} 
    \State Add each $(x, z_t, z_{t+1}, s)$ tuple to replay buffer
\Else
    \State Add $(x, s)$ tuple to replay buffer
\EndIf 
\end{algorithmic}
\end{algorithm}

\begin{algorithm}
\caption{E-step Inner Loop: Dynamics}\label{alg:attractors:e_step_dynamics}
\begin{algorithmic}[1]

\Require  $x \in \mathcal{X}, z_t \in \mathcal{Z}, z_{t+1} \in \mathcal{Z}, s \in \mathcal{S}$ \Comment From replay buffer
\Statex
\State \texttt{logpz\_fw} $\leftarrow \log P_\theta(z_{t+1} \mid z_t)$ \Comment Forward dynamics 
\State \texttt{logpz\_bw} $\leftarrow \log P^\leftarrow_\theta(z_t \mid z_{t+1}, t+1, x)$ \Comment Backward dynamics
\Statex
\State \texttt{logF1} $\leftarrow \log F_\theta(z_t, s, t, x)$ \Comment Compute flow (Alg.~\ref{alg:attractors:flow})
\State \texttt{logF2} $\leftarrow \log F_\theta(z_{t+1}, s, t+1, x)$ \Comment Compute flow (Alg.~\ref{alg:attractors:flow})
\Statex
\State $\mathcal{L} \leftarrow \left ( \texttt{logF1} + \texttt{logpz\_fw} - \texttt{logF2} - \texttt{logpz\_bw} \right )^2$
\State Update $\theta$ using $\nabla \mathcal{L}$

\end{algorithmic}
\end{algorithm}

\begin{algorithm}
\caption{Compute Flow}\label{alg:attractors:flow}
\begin{algorithmic}[1]
\Require $z \in \mathcal{Z}, s \in \mathcal{S}, t \geq 0, x \in \mathcal{X}$
\Statex
\State \texttt{logPs\_fw} $\leftarrow \log P_\theta(s_0 \mid z) + \sum\limits_{l=0}^{L-1} \log P_\theta(s_{l+1} \mid s_0, \ldots, s_l, z)$
\State \texttt{logPs\_bw} $\leftarrow \log P_\theta(z \mid \hat z_s) + \sum\limits_{l=0}^{L-1} \log P_\theta(s_l \mid s_0, \ldots, s_{l+1}, z)$
\State \texttt{logR} $\leftarrow \log R_\phi(s, x)$ 
\State \texttt{g} $\leftarrow g_\theta(z, t, x)$
\State \texttt{logF} $\leftarrow$ \texttt{logR} + \texttt{logPs\_bw} - \texttt{logPs\_fw} + \texttt{g}
\Statex
\State return \texttt{logF}
\end{algorithmic}
\end{algorithm}

\begin{algorithm}
\caption{E-step Inner Loop: Discretizer}\label{alg:attractors:e_step_discretizer}
\begin{algorithmic}[1]
\Require $x \in \mathcal{X}, z \in \mathcal{Z}$ \Comment From replay buffer
\Statex
\State Sample $s \sim P_\theta(s \mid z)$
\Statex
\State \texttt{logPs\_fw} $\leftarrow \log P_\theta(s_0 \mid z) + \sum\limits_{l=0}^{L-1} \log P_\theta(s_{l+1} \mid s_0, \ldots, s_l, z)$
\State \texttt{logPs\_bw} $\leftarrow \log P_\theta(z \mid \hat z_s) + \sum\limits_{l=0}^{L-1} \log P_\theta(s_l \mid s_0, \ldots, s_{l+1}, z)$
\State \texttt{logR} $\leftarrow \log R_\phi(s, x)$ 
\State \texttt{logZ} $\leftarrow \log Z_\theta(z, x)$
\Statex
\State $\mathcal{L} \leftarrow \left ( \texttt{logZ} + \texttt{logPs\_fw} - \texttt{logR} - \texttt{logPs\_bw} \right )^2$
\State Update $\theta$ using $\nabla \mathcal{L}$

\end{algorithmic}
\end{algorithm}

\begin{algorithm}
\caption{M-step Inner Loop}\label{alg:attractors:m_step}
\begin{algorithmic}[1]
\Require $x \in \mathcal{X}, s \in \mathcal{S}$ \Comment From replay buffer
\Statex
\State Sample $z_0 \sim P_\phi(z_0 \mid x)$
\Statex
\State \texttt{recon\_error} $\leftarrow$ $-\log P_\phi(x, z_0)$
\State \texttt{kl\_prior} $\leftarrow D_{\rm KL}( P_\phi(z_0 \mid x)\ \|\ P(z_0))$
\State \texttt{info\_loss} $\leftarrow D_{\rm KL}( P_\phi(z_0 \mid x)\ \|\ P^{SG}_\phi(z_0 \mid \hat z_s)) + \beta D_{\rm KL}( P^{SG}_\phi(z_0 \mid x)\ \|\ P_\phi(z_0 \mid \hat z_s))$
\Statex
\State $\mathcal{L}$ = \texttt{recon\_error} + \texttt{kl\_prior} + \texttt{info\_loss}
\State Update $\phi$ using $\nabla \mathcal{L}$
\end{algorithmic}
\end{algorithm}
\end{document}